%% file: main.tex
\documentclass[accepted]{uai2026} 
                        
\usepackage[american]{babel}

\usepackage{natbib} 
\usepackage{mathtools} 
\usepackage{booktabs} 
\usepackage{tikz} 
\usepackage{graphicx}
\usepackage{subfig}
\usepackage{caption}
\usepackage{subcaption}
\usepackage{amssymb}
\usepackage{algorithm}
\usepackage[noend]{algorithmic}

\DeclareMathOperator*{\argmax}{arg\,max}

\usepackage{xcolor}

\definecolor{oxfordblue}{RGB}{4,30,66}

\input{macros.tex}

\title{Verbalizing LLM’s Higher-order Uncertainty via Imprecise Probabilities}

\author[1,2]{Anita Yang}
\author[3]{Krikamol Muandet}
\author[4,5]{Michele Caprio}
\author[6,*]{Siu Lun Chau}
\author[1,*]{\href{mailto:<masaki_adachi@mail.toyota.co.jp>?Subject=Your UAI 2026 paper}{Masaki Adachi}}
\affil[1]{%
    Lattice Lab\\
    Toyota Motor Corporation\\ Japan
}
\affil[2]{%
    Department of Computer Science\\
    University of Tokyo\\
    Japan
}
\affil[3]{%
    Rational Intelligence Lab\\
    CISPA\\
    Helmholtz Center for Information Security\\
    Germany
}
\affil[4]{%
    Department of Computer Science\\
    University of Manchester\\
    United Kingdom
}
\affil[5]{%
    Manchester Centre for AI Fundamentals\\
    United Kingdom
}
\affil[6]{%
    EPIC Lab\\
    College of Computing \& Data Science\\
    Nanyang Technological University\\
    Singapore
}
\affil[*]{%
    Equal contribution
}
  
\begin{document}
\maketitle

\begin{abstract}
    Despite the growing demand for eliciting uncertainty from large language models (LLMs), empirical evidence suggests that LLM behavior is not always adequately captured by the elicitation techniques developed under the classical probabilistic uncertainty framework.
    This mismatch leads to systematic failure modes, particularly in settings that involve ambiguous question-answering, in-context learning, and self-reflection.
    To address this, we propose novel prompt-based uncertainty elicitation techniques grounded in \emph{imprecise probabilities}, a principled framework for representing and eliciting higher-order uncertainty.
    Here, first-order uncertainty captures uncertainty over possible responses to a prompt, while second-order uncertainty (uncertainty about uncertainty) quantifies indeterminacy in the underlying probability model itself.
    We introduce general-purpose prompting and post-processing procedures to directly elicit and quantify both orders of uncertainty, and demonstrate their effectiveness across diverse settings.
    Our approach enables more faithful uncertainty reporting from LLMs, improving credibility and supporting downstream decision-making.
\end{abstract}

\input{sections/01_introduction}

\input{sections/02_method}

\input{sections/03_synthetic_exp}

\input{sections/04_realworld_exp}

\input{sections/05_conclusion}





\bibliography{reference}

\newpage

\onecolumn

\title{Verbalizing LLM’s Higher-order Uncertainty via Imprecise Probabilities\\(Supplementary Material)}
\maketitle

\appendix

\vspace{20pt}
\section{Prompts}
\subsection{Second-Order Uncertainty}
We provide prompts for eliciting two other IP representations of second-order uncertainty discussed in the main text: credal sets (\textsc{Credal}) and possibility functions (\textsc{Pos}).

\textbf{Credal sets.}
We represent the credal set by eliciting a finite ensemble of $M$ precise predictive distributions 
$\smash{\{p^{(m)}\}_{m=1}^M}$ from $M$ models or samples (random seed). The resulting credal set is taken to be the empirical set of these distributions: $\smash{\mathcal{C} = \mathrm{conv}(\{p^{(m)}\}_{m=1}^M)}$. Each $\smash{p^{(m)}}$ is obtained using Prompt~\ref{prompt:credal}.

\begin{figure}[h]
\centering
\begin{promptbox}[Credal]{prompt:credal}
Assign a probability (between 0.0 and 1.0) representing how likely it is that the answer would be given as a response to the question.\\
A correct answer should generally receive a higher probability than an incorrect one. Likelihood may vary based on reasonable interpretations of the question (e.g., ambiguity in scope, answer type, entity interpretation, or contextual assumptions).\\
The sum of all probabilities must not exceed 1.0.
\end{promptbox}
\end{figure}

\textbf{Possibility function.}
We elicit the LLM to provide a possibility function over the candidate answers, including an explicit alternative option (i.e., ``none of the above''), using Prompt~\ref{prompt:possibility}. We allow the elicited scores to be unnormalized; the required normalization is applied as part of the MMI computation (Apd.~\ref{apd:mmi_posfunc}).

\begin{figure}[h]
\centering
\begin{promptbox}[Pos]{prompt:possibility}
Provide a possibility score which captures how plausible the answer correctly answers the question.\\
Then, provide a possibility score how plausible it is that a different answer (not listed) could be correct.\\
The possibility should be between 0.0 and 1.0. A possibility score of 1.0 means ``fully plausible,'' and 0.0 means ``impossible.''
\end{promptbox}
\end{figure}

\subsection{Approximation of $\hat{\mathcal{Y}}$.} \label{apd:approx_Yhat}
To characterize uncertainty about the question $x$ itself---rather than about a particular answer $y$ (e.g., a predicted $\hat{y}$)---we consider uncertainty over the candidate answer set $\mathcal{Y}$ (Apd.~\ref{apd:correctness_vs_Y}). In open-ended QA, however, $\mathcal{Y}$ is not observed. We therefore approximate it with a finite set $\hat{\mathcal{Y}}$ by prompting the model to generate plausible (non-zero-probability) candidate answers (Prompt~\ref{prompt:candidates}), within which the ground-truth set of correct answers $\mathcal{Y}^\star$ is likely to lie.

\begin{figure}[h]
  \centering
\begin{promptbox}[Candidate answers $\hat{\mathcal{Y}}$]{prompt:candidates}
Given the question below, generate a list of all possible correct answers, taking into account different reasonable interpretations of the question.\\

Provide the answers as a numbered list, with each answer on its own line.
Each answer must be concise text only, with no explanations or additional wording.
Do not include duplicates or answers that refer to the same entity or concept.
For example:\\
1. <answer one as concise text>\\
2. <answer two as concise text>\\
...
\end{promptbox}
\end{figure}

\section{Maximum Mean Imprecision} \label{apd:mmi}

\begin{figure}[h]
    \centering
    \begin{minipage}[t]{0.45\linewidth}
        \centering
        \includegraphics[width=0.7\linewidth]{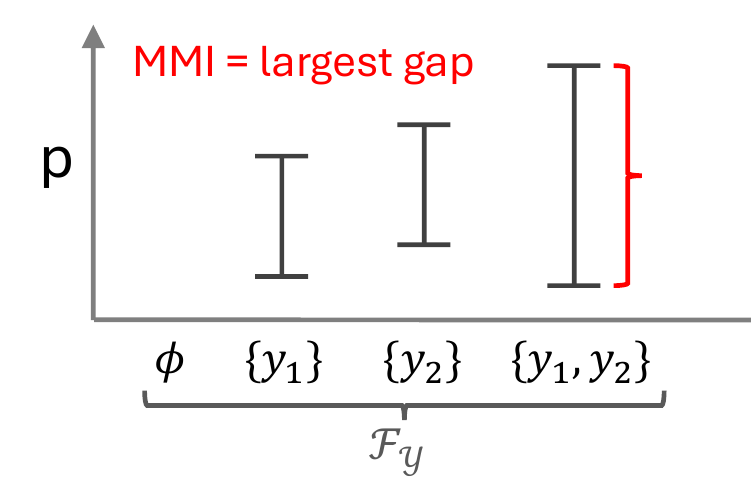}
        \caption{MMI measures the largest gap between upper/lower probabilities across all events $\smash{A \in \mathcal{F_Y}}$.}
        \label{fig:mmi_exact_vis}
    \end{minipage}\hspace{10pt}
    \begin{minipage}[t]{0.45\linewidth}
        \centering
        \includegraphics[width=0.65\linewidth]{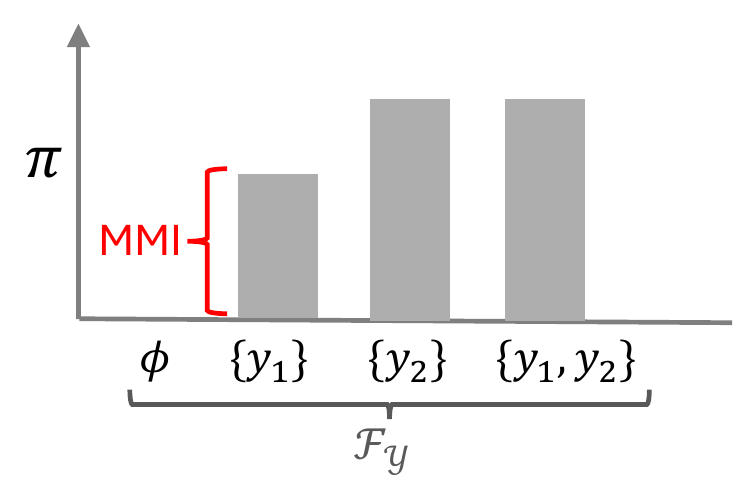}
        \caption{MMI for possibility function measures the second largest possibility score over $\mathcal{Y}$. }
        \label{fig:pos_mmi}
    \end{minipage}
\end{figure}

The exact MMI metric for measuring EU~\citep{chau2025integral} under total variation is
\begin{align}
  \text{MMI}_{\text{TV}}(\underline{P})
  &:= \sup_{A \in \mathcal{F}_{\mathcal{Y}}} \left(\overline{P}(A) - \underline{P}(A)\right)
  \;\le\; 1 - \sum_{y \in \mathcal{Y}} \underline{P}(\{y\}),\label{eq:mmi}
\end{align}
where $\mathcal{F}_{\mathcal{Y}}$ is the $\sigma$-algebra over the candidate answers $\mathcal{Y}$ (in our discrete setting, $\mathcal{F}_{\mathcal{Y}} = 2^{\mathcal{Y}}$). The lower and upper probabilities $\underline{P}(A)$ and $\overline{P}(A)$ bound the probability of an event $A \subseteq \mathcal{Y}$, which we interpret as “the sampled answer lies in $A$.” The quantity $\overline{P}(A)-\underline{P}(A)$ is the imprecision (interval width) for event $A$, and $\text{MMI}_{\text{TV}}$ takes the maximum such width over all events. For example, when $\mathcal{Y}=\{y_1,y_2\}$, Fig.~\ref{fig:mmi_exact_vis} enumerates all events in $2^{\mathcal{Y}}$ and their bounds; $\text{MMI}_{\text{TV}}$ corresponds to the widest interval among them, capturing the agent’s worst-case epistemic uncertainty over the output space.

\subsection{Credal set}
In \textsc{Credal-EU}, we instantiate the credal set as the convex hull of a finite ensemble of $M$ precise predictive distributions $\{p^{(m)}\}_{m=1}^M$ produced by independently prompted runs (Prompt~\ref{prompt:credal}):
\begin{equation}
  \mathcal{C} := \mathrm{conv}\big(\{p^{(m)}\}_{m=1}^M\big).
\end{equation}
Since $\sum_{y\in A} p(y)$ is linear in $p$, its extrema over the polytope $\mathcal{C}$ are attained at extreme points. Therefore, the event bounds reduce to simple min/max over ensemble members:
\begin{equation}
  \underline{P}(A)=\min_{m \in [M]} \sum_{y \in A} p^{(m)}(y),
  \qquad
  \overline{P}(A)=\max_{m \in [M]} \sum_{y \in A} p^{(m)}(y).
  \label{eq:credal_event_bounds}
\end{equation}

In particular, the singleton bounds used in the tractable upper bound are
\begin{equation}
  \underline{P}(\{y\})=\min_{m \in [M]} p^{(m)}(y),
  \qquad
  \overline{P}(\{y\})=\max_{m \in [M]} p^{(m)}(y).
  \label{eq:credal_singleton_bounds}
\end{equation}
Exact $\text{MMI}_{\text{TV}}$ requires optimizing over events $A$ and can be exponential in $|\mathcal{Y}|$. In our experiments, we therefore (i) compute exact MMI only for \emph{singleton} events corresponding to a specific answer $y$ (which reduces to the interval width in Eq.~\ref{eq:credal_singleton_bounds}), and (ii) use the linear-time upper bound in Eq.~\ref{eq:mmi} for uncertainty over $\mathcal{Y}$. For open-ended QA, where $\mathcal{Y}$ is approximated by a finite candidate set $\hat{\mathcal{Y}}$, we apply the same computations over $\hat{\mathcal{Y}}$.

\subsection{Probability Interval}
In \textsc{ProbInt-EU}, the LLM elicits a probability interval $[\underline{p}(y), \overline{p}(y)]$ for each candidate answer $y$. For singleton events, the induced lower/upper probabilities are immediate:
\begin{equation}
  \underline{P}(\{y\})=\underline{p}(y),
  \qquad
  \overline{P}(\{y\})=\overline{p}(y).
  \label{eq:probint_singleton_bounds}
\end{equation}
For non-singleton events $A \subseteq \mathcal{Y}$, constructing coherent bounds $\underline{P}(A)$ and $\overline{P}(A)$ from singleton intervals generally requires additional assumptions  and can be more involved. Moreover, computing exact $\text{MMI}_{\text{TV}}$ still entails optimizing over events $A$ and can scale exponentially with $|\mathcal{Y}|$. We therefore follow the same strategy as \textsc{Credal-EU}: (i) we compute \emph{exact} MMI only for answer-level uncertainty on a specific $y$ (i.e., singleton events / correctness of $y$), and (ii) we use the linear-time upper bound in Eq.~\ref{eq:mmi} for task-level uncertainty over $\mathcal{Y}$ (or its approximation $\hat{\mathcal{Y}}$).

\subsection{Possibility function}\label{apd:mmi_posfunc}
In \textsc{Pos-EU}, we prompt the LLM to elicit a (potentially unnormalized) possibility function $\hat{\pi}:\mathcal{Y}\to[0,1]$ over the output space. In open-ended QA, $\mathcal{Y}$ is approximated by a finite candidate set $\hat{\mathcal{Y}}$ together with a ``none of the above'' option, allowing the model to assign high plausibility to an outcome outside the enumerated candidates without forcing plausibility to be redistributed among them, since possibility scores are non-additive.

For possibility functions, total-variation MMI admits a simple form: after normalization so that $\max_{y\in\mathcal{Y}}\pi(y)=1$, MMI equals the second-largest possibility value~\citep{chau2026quantifying}:
\begin{align}
    \text{MMI}_\text{TV} &= \pi_{(2)} \\
    &= \frac{\hat{\pi}_{(2)}}{\hat{\pi}_{(1)}}, \label{eq:mmi_pos}
\end{align}
where $\pi_{(1)}\ge \pi_{(2)}\ge \cdots$ denote the order statistics of the normalized scores $\{\pi(y)\}_{y\in\mathcal{Y}}$, and $\hat{\pi}_{(1)}\ge \hat{\pi}_{(2)}\ge \cdots$ are the corresponding order statistics of the unnormalized scores $\{\hat{\pi}(y)\}$. Because normalization enforces $\pi_{(1)}=1$, $\pi_{(2)}$ measures the strength of the best competing alternative relative to the top hypothesis: larger values indicate greater ambiguity among leading candidates (and hence higher imprecision). Equivalently, MMI quantifies uncertainty in the top answer via the strongest competitor (Fig.~\ref{fig:pos_mmi}).

When assessing uncertainty for a single answer $y$ (binary event $\{y,\neg y\}$), we elicit unnormalized scores $\hat{\pi}(y)$ and $\hat{\pi}(\neg y)$ (with $\neg y$ implemented as ``not $y$'') and normalize by dividing by $\max\{\hat{\pi}(y),\hat{\pi}(\neg y)\}$, resulting in:
\begin{equation}
\text{MMI}_\text{TV}
=\min\!\left\{\frac{\hat{\pi}(y)}{\max(\hat{\pi}(y),\hat{\pi}(\neg y))},\ 
\frac{\hat{\pi}(\neg y)}{\max(\hat{\pi}(y),\hat{\pi}(\neg y))}\right\}
=\frac{\min\{\hat{\pi}(y),\hat{\pi}(\neg y)\}}{\max\{\hat{\pi}(y),\hat{\pi}(\neg y)\}}.
\end{equation}

\subsection{Approximate vs. Exact} \label{apd:approx_vs_exact}
We empirically compare the approximate against the exact MMI  (Eq.~\ref{eq:mmi}). On AmbigQA$^*$ and MAQA$^*$, the Pearson correlations between the approximate and exact estimates are $0.9318$ and $0.9266$, respectively, indicating strong agreement. This agreement comes with a substantial computational advantage: the exact MMI computation scales exponentially with the number of candidate answers $|\mathcal{Y}|$, whereas the approximation scales linearly in $|\mathcal{Y}|$. For example, with $|\mathcal{Y}|=20$ over 200 AmbigQA$^*$ samples, exact MMI computation takes 627.2 seconds, while the approximation takes less than 0.0001 seconds.

\section{Uncertainty on: ``answer $y$'' vs. ``candidate answers $\mathcal{Y}$''} \label{apd:correctness_vs_Y}
We clarify when uncertainty should be defined for the correctness of \emph{a particular answer $y$} (e.g., a model prediction $\hat{y}$) versus over the \emph{candidate answer set} $\mathcal{Y}$ (or its approximation $\hat{\mathcal{Y}}$).

\textbf{Two objects of uncertainty:}\\
(i) \emph{Answer $y$ (answer-centric).} Uncertainty on $y$ concerns whether a particular answer $y$ (typically $\hat{y}$) is correct, collapsing all alternatives into ``not-$y$.'' It is most useful for decision-focused use cases (trust/abstain/verify/fallback), where utility depends primarily on the correctness of the chosen answer.

(ii) \emph{Candidate answers $\mathcal{Y}$ (question-centric).} Uncertainty on $\mathcal{Y}$ concerns how belief is distributed across plausible answers. It is most useful for question-focused use cases (ambiguity detection, multiple valid answers, need for clarification), where utility depends on whether the question supports several competing answers, not just whether $\hat{y}$ is correct.

\subsection{First-order uncertainty}

\textbf{On answer $y$.}
For a given answer $y$, first-order uncertainty is defined on the binary event ``$y$ vs.\ not-$y$.'' Under a precise predictive distribution $p$, this is the Bernoulli entropy $H\!\left(\mathrm{Bern}(p(y))\right)$. In Fig.~\ref{fig:au_correctness}, the correctness AU for $y_1$ equals $H(\mathrm{Bern}(0.4))$.

\textbf{On candidate answers $\mathcal{Y}$.}
First-order uncertainty over the candidate set quantifies how probability mass is distributed \emph{across} answers. Under a precise predictive distribution $p$ on $\mathcal{Y}$, this is the entropy $H(p)$ over all $y\in\mathcal{Y}$ (Fig.~\ref{fig:au_correctness}).

\textbf{Experiments.}
We use these notions for different evaluation goals:
(i) \emph{Answer $y$}: decision-centric evaluations, including error tracking in our ICL experiments (Sec.~\ref{sec:synthetic}) and correctness detection (AUROC; Fig.~\ref{fig:auroc_total} in Sec.~\ref{sec:real-world}).
(ii) \emph{Candidate answers $\mathcal{Y}$}: question-centric evaluations, namely ambiguity detection in open-ended QA (Fig.~\ref{fig:au_auroc} in Sec.~\ref{sec:real-world}).

\subsection{Second-order uncertainty}

\begin{figure}[h]
    \centering
    \subfloat[First order: (i) binary entropy for $y$ vs.\ not-$y$, or (ii) entropy over $\mathcal{Y}$.\label{fig:au_correctness}]{%
        \includegraphics[width=0.23\linewidth]{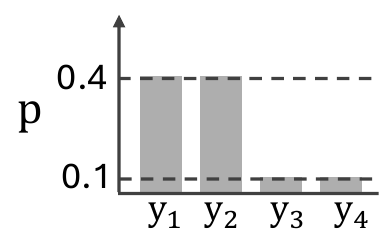}
    }\hspace{40pt}
    \subfloat[Second order: (i) interval width of each $y$, or (ii) upper-bound MMI over $\mathcal{Y}$.\label{fig:eu_correctness}]{%
        \includegraphics[width=0.23\linewidth]{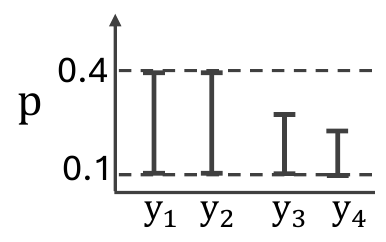}
    }
    \caption{First- and second-order uncertainty can be computed either for the (i) correctness of a specific answer $y$, or (ii) over the candidate answers set $\mathcal{Y}$}
    \label{fig:correctness_vs_answers}
\end{figure}

\textbf{On answer $y$.}
For a given answer $y$, second-order uncertainty (imprecision) captures uncertainty about the binary event ``$y$ vs.\ not-$y$'' under a probability interval. We measure this by the interval width $\overline{p}(y)-\underline{p}(y)$ (Eq.~\ref{eq:mmi_binary}). In Fig.~\ref{fig:eu_correctness}, the EU for $y_2$ is the interval gap (MMI $=0.3$).

\textbf{On candidate answers $\mathcal{Y}$.}
Second-order uncertainty over the candidate set captures imprecision about the \emph{entire} answer distribution. We measure this using the upper-bound MMI,
$1-\sum_{y\in\mathcal{Y}}\underline{p}(y)$ (Eq.~\ref{eq:mmi_upper}). In Fig.~\ref{fig:eu_correctness}, this aggregates lower bounds across candidates (upper-bound MMI $=1-0.4$).

\textbf{Experiments.}
We use these notions for different evaluation goals:
(i) \emph{Answer $y$}: decision-centric evaluations, including error tracking in our ICL experiments (Sec.~\ref{sec:synthetic}) and correctness detection (AUROC; Sec.~\ref{sec:real-world}).
(ii) \emph{Candidate answers $\mathcal{Y}$}: question-centric evaluations, including ambiguity-related analyses and comparisons to ``ground-truth'' EU distributions (Fig.~\ref{fig:eu_with_au1}).

\section{Experiments}

\subsection{In-context learning} \label{apd:icl}

\subsubsection{Second-order denoise}
\begin{figure}[h]
    \centering
    \subfloat[\textsc{Pos}\label{fig:eu_with_au_icl_pos_direct}]{%
        \includegraphics[width=0.38\linewidth]{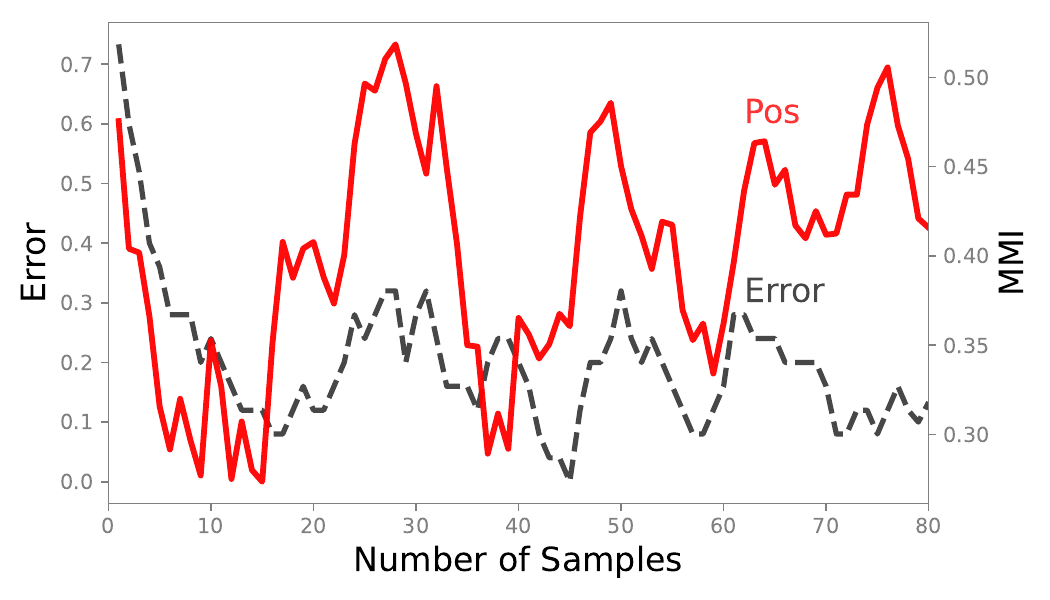}
    }\hspace{20pt}
    \subfloat[\textsc{Credal}\label{fig:eu_with_au_icl_credal}]{%
        \includegraphics[width=0.38\linewidth]{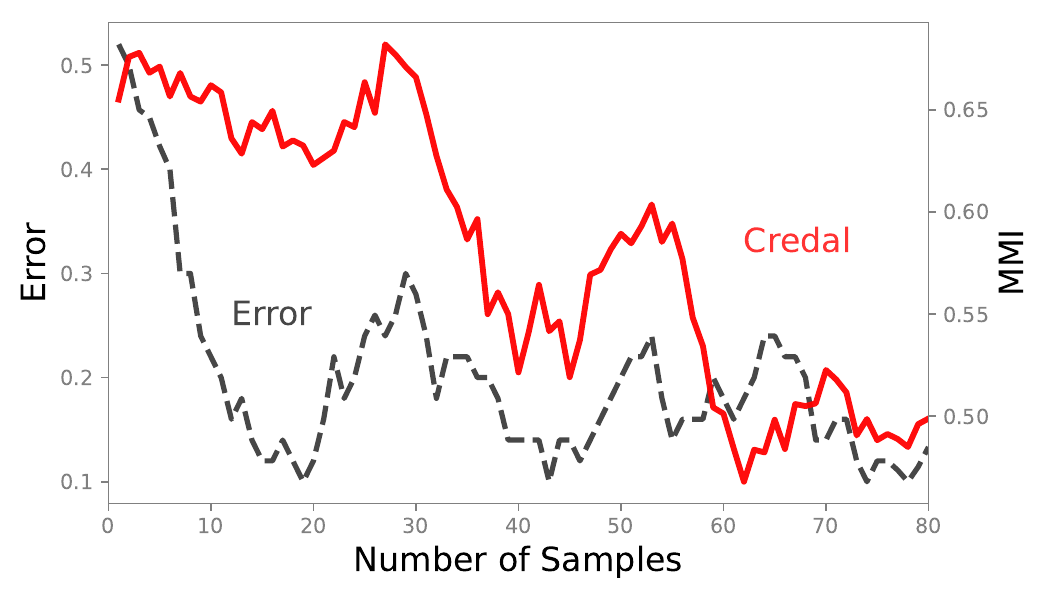}
    }
    \caption{Approximations of second-order uncertainty with fixed first-order noise (lower-case probability $p=0.25$) (with moving average window of 10). Both \textsc{Pos} and \textsc{Credal} generally match shape of the error.}
    \label{fig:eu_with_au_icl_both}
\end{figure}

We report additional error-tracking results from Sec.~\ref{sec:icl_eu} using second-order uncertainty approximated via \textsc{Credal} (Fig.~\ref{fig:eu_with_au_icl_credal}) and \textsc{Pos} (Fig.~\ref{fig:eu_with_au_icl_pos_direct}). In contrast to \textsc{Vanilla} (Fig.~\ref{fig:au_fixed_eu}), both representations track error more closely, and \textsc{Credal} shows a clear decrease as the number of in-context examples grows, indicating effective denoising of epistemic uncertainty with additional evidence. Between \textsc{ProbInt}, \textsc{Pos}, and \textsc{Credal}, \textsc{Pos} is the most sensitive to first-order noise, consistent with Fig.~\ref{fig:au_noise} (holding second-order fixed while varying first-order noise). Nevertheless, \textsc{Pos} remains more robust than \textsc{Vanilla} under first-order noise (Fig.~\ref{fig:au_noise}).

\subsubsection{Group Uncertainty Elicitation}\label{apd:icl_ensemble}

We provide additional experimental details in Sec.~\ref{sec:icl_ensemble}. Our ensemble consists of three language models: GPT-5-Mini, GPT-4.1-mini, and GPT-5-Nano. Using GPT-5-Mini with Prompt~\ref{prompt:candidates}, we first construct a shared candidate set $\hat{\mathcal{Y}}$. This controls for the fact that different models may produce different point predictions $\hat{y}$, while our elicitation procedure requires every model to assign uncertainty over the same set of answer options. We therefore define $\hat{\mathcal{Y}}$ to cover the answers the ensemble is most likely to generate, and then prompt each model to verbalize its uncertainty over $\hat{\mathcal{Y}}$. In contrast, for our other \textsc{Credal} experiments we obtain multiple samples by repeatedly querying a single model.

\subsubsection{Prompt sensitivity}\label{apd:prompt_sensitivity}
\begin{figure}[!h]
    \centering
    \includegraphics[width=0.5\linewidth]{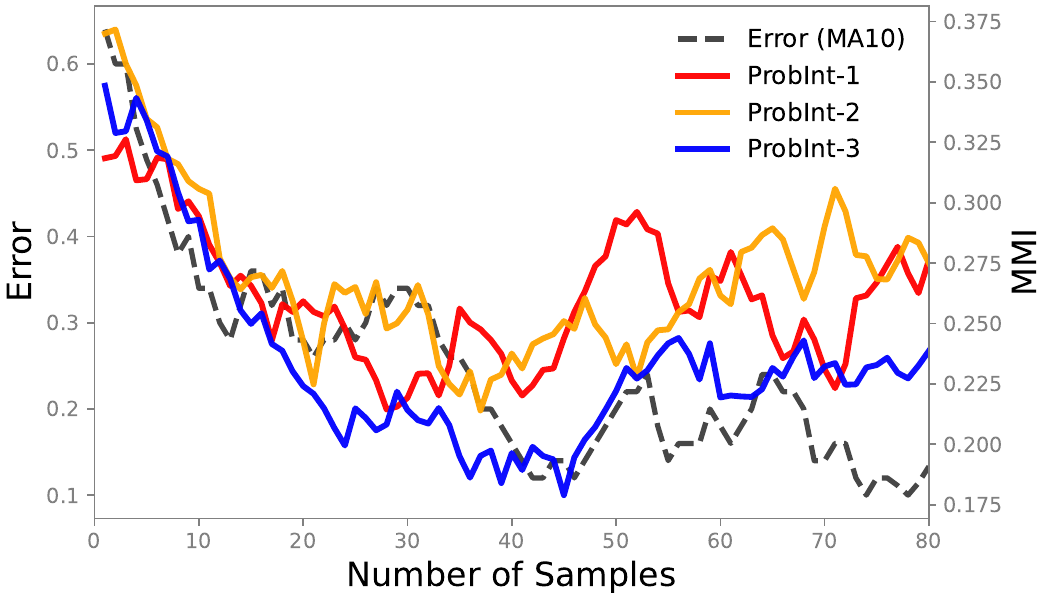}
    \caption{Second-order uncertainty estimates elicited by variants of the \textsc{ProbInt} prompt consistently track the overall error trend.}
    \label{fig:probint_variants}
\end{figure}

We analyze the sensitivity of our method to prompt formulation. Specifically, we consider three variants, denoted \textsc{ProbInt-1}--\textsc{ProbInt-3} (Prompts~\ref{prompt:probint1}--\ref{prompt:probint3}). We evaluate these prompts on the synthetic in-context learning setup from Sec.~\ref{sec:synthetic}. As shown in Fig.~\ref{fig:probint_variants}, the second-order uncertainty elicited by each prompt qualitatively tracks the error trend, suggesting that the method is robust across prompt formulations.

\begin{figure}[h]
\centering
\begin{promptbox}[ProbInt-1]{prompt:probint1}
Provide your best guess to the question and a lower and upper probability (each between 0.0 and 1.0) that your guess is correct.\\

Interpret the probabilities as follows:\\
- Lower Probability: the smallest probability you consider plausible that your guess is correct.\\
- Upper Probability: the largest probability you consider defensible that your guess is correct.
\end{promptbox}
\end{figure}

\begin{figure}[h]
\centering
\begin{promptbox}[ProbInt-2]{prompt:probint2}
An unknown transformation maps each input sequence to an output sequence.\\
Now predict the output for the given input and state your uncertainty as an upper and lower probability (each between 0.0 and 1.0) that your answer is correct.\\

Interpret the probabilities as follows:\\
- Upper Probability: the largest probability you consider defensible that your output is correct.\\
- Lower Probability: the smallest probability you consider plausible that your output is correct
\end{promptbox}
\end{figure}

\begin{figure}[h]
\centering
\begin{promptbox}[ProbInt-3]{prompt:probint3}
The following input-output pairs were produced by an unknown sequence transformation.\\
Based only on these examples, predict the output sequence for the test input below. Then give a probability interval for the event that your predicted output is correct.\\

Interpretation:\\
- Lower Probability: the conservative plausible probability that your output is correct.\\
- Upper Probability: the highest defensible probability that your output is correct.
\end{promptbox}
\end{figure}

\subsection{Real-World QA Experiment}

\subsubsection{Estimating first- and second-order uncertainty from dataset}\label{apd:real}
We assess disentangling and quantifying second-order uncertainty in settings where both first- and second-order uncertainty are present. In such cases, AUROC can be reductive: it scores correctness against a single reference answer $y^\star$, even when other answers may also be valid (with varying probabilities). To enable a more principled evaluation, we follow \citet{tomov2025illusion}, who propose an approximation to \emph{ground-truth} first- and second-order uncertainty distributions (their aleatoric uncertainty (AU) and epistemic uncertainty (EU)) for ill-defined real-world QA. They posit a \emph{true} first-order answer distribution $p^\star$, which can be approximated from corpus statistics by estimating question--answer co-occurrence frequencies in a large text corpus. This estimated $p^\star$ is provided in the AmbigQA$^\star$ and MAQA$^\star$ datasets.

Given $p^\star$ and a model predictive distribution $\hat{p}$, AU and EU admit the decomposition
\begin{align}
    \underbrace{\mathrm{CE}(p^\star, \hat{p})}_\text{Total uncertainty} = \underbrace{H(p^\star)}_\text{AU} + \underbrace{\mathrm{KL}(p^\star \, \| \, \hat{p})}_\text{EU}, \label{eq:eu}
\end{align}
where $\mathrm{CE}$ denotes cross-entropy and $\mathrm{KL}$ the Kullback--Leibler divergence. This cross-entropy decomposition is closely related to mutual-information-based uncertainty decompositions~\citep{kotelevskii2025from}. We therefore treat $H(p^\star)$ as a proxy for ground-truth first-order uncertainty (AU) and $\mathrm{KL}(p^\star \,\|\, \hat{p})$ as a proxy for ground-truth second-order uncertainty (EU).

\textbf{Metric.} Following \citet{tomov2025illusion}, we use concordance statistics~\citep{therneau2023concordance} to measure rank agreement between an estimated uncertainty score and the corresponding proxy ground truth. Concordance is the probability that, for a randomly chosen pair of examples, the method assigns a higher uncertainty to the example with higher ground-truth uncertainty. Its interpretation matches AUC-ROC: values closer to 1 indicate better ranking alignment. We report this metric in Fig.~\ref{fig:eu_with_au1}.

\textbf{Second-order uncertainty.}
We compare second-order uncertainty estimated by our proposed IP representations to the KL-based proxy $\mathrm{KL}(p^\star \,\|\, \hat{p})$. Figure~\ref{fig:maqa_eu_with_au} visualizes their association for \textsc{Vanilla}, \textsc{CoT}, and \textsc{ProbInt}; our method exhibits the strongest alignment. We further benchmark against additional uncertainty estimators in Fig.~\ref{fig:eu_with_au1} using the concordance index, where our methods are the most aligned across datasets and models.

\begin{figure}[h]
    \centering
    \includegraphics[width=0.6\linewidth]{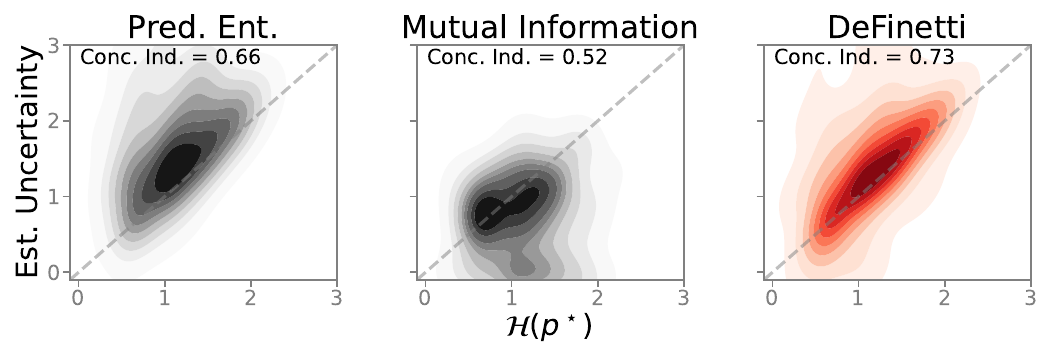}
    \caption{\textsc{DeFinetti} best aligns with the proxy for first-order uncertainty}
    \label{fig:au_ci_vis}
\end{figure}

\paragraph{First-order uncertainty.} We also evaluate first-order uncertainty by comparing method estimates to the entropy proxy $H(p^\star)$ on MAQA$^\star$. Figure~\ref{fig:au_ci_vis} visualizes the estimated first-order uncertainty from \textsc{Predictive Entropy}, \textsc{MI-Clarifications}, and \textsc{DeFinetti}, where \textsc{DeFinetti} is best aligned with the proxy with highest concordance index.

\subsubsection{Consistency of second-order with first-order elicitation} \label{apd:consistency}
We evaluate whether the first-order point estimate elicited by Prompt~\ref{prompt:definetti} lies within the probability interval elicited by Prompt~\ref{prompt:interval} on AmbigQA under the setting of Fig.~\ref{fig:auroc_total}, where both types of uncertainty are required. With the original prompts, the containment ratio is $72.33\%$, with an AUROC of $0.7862 \pm 0.0035$. Adding an explicit containment constraint raises the containment ratio to $100\%$, while AUROC remains comparable at $0.7554 \pm 0.0008$. Thus, the qualitative conclusion remains unchanged.

\subsubsection{Smaller LLMs}
We provide an additional evaluation with Llama-3.1-8B on the Non-AmbigQA experiment in Table~\ref{tab:aleatoric_uncertainty_gpt5_question}, where the model achieves $79\%$ prediction accuracy. We compare \textsc{ProbInt} against \textsc{Label Prob.}, which prompts the model five times and estimates confidence from the frequency of the predicted answer. However, \textsc{Label Prob.} produces the same answer across samples, resulting in an AUROC of $0.5$. In contrast, \textsc{ProbInt} achieves an AUROC of $0.6027 \pm 0.0000$, outperforming the \textsc{Vanilla} baseline at $0.5698 \pm 0.0017$. We also attempted to extend the evaluation to other smaller models, but found that they generally lack sufficient capacity to solve the underlying task, making uncertainty tracking uninformative for all methods.

\end{document}

%% file: macros.tex
\usepackage[most]{tcolorbox}
\tcbuselibrary{listings, skins, breakable}

\newcounter{prompt}

\newtcolorbox[use counter=prompt]{promptbox}[2][]{%
  colback=gray!5,
  colframe=gray!30,
  fonttitle=\bfseries,
  coltitle=black,
  boxrule=0.4pt,
  arc=2pt,
  left=6pt,
  right=6pt,
  top=6pt,
  bottom=6pt,
  width=\linewidth,
  enhanced,
  title={Prompt~\thetcbcounter: #1},
  label={#2},
}

%% file: sections/01_introduction.tex
\section{Introduction}\label{sec:intro}
\begin{figure}[t]
  \centering
  \includegraphics[width=1\linewidth]{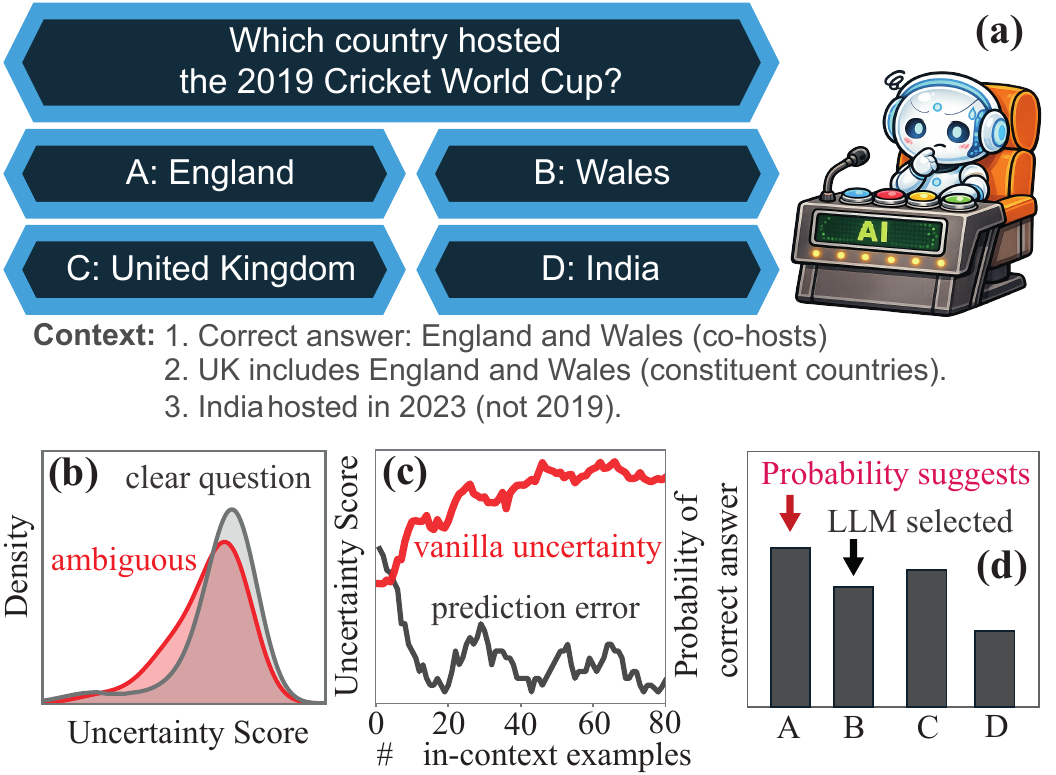}
  \caption{
  Collection of failure modes in prior verbalized uncertainty scores.
  (a) Example of an ambiguous question.
  (b) The prior uncertainty score fails to distinguish between clear and ambiguous question distributions.
  (c) The prior score also fails to track the decrease in prediction error, which should reflect reduced uncertainty as more in-context examples are provided.
  (d) Self-reflection on answer-wise probability/uncertainty should explain the rationale behind answer selection, but it often fails to do so.
  }
  \label{fig:concept}
\end{figure}

Uncertainty quantification~(UQ) for large language models (LLMs)~\citep{shorinwa2025surveyonuq} has been proven effective across many downstream tasks, including hallucination detection~\citep{bouchard2025uncertainty, farquhar2024detecting, tomani2024uncertainty}, reasoning enhancement~\citep{lugoloobi2026LLMs}, active learning~\citep{wang2024active}, model selection~\citep{agrawal2025uncertainty}, and agentic workflow control~\citep{tomani2024uncertainty, machcha2025large}. 
Because most state-of-the-art LLMs are closed-source, research on uncertainty elicitation has largely focused on \emph{verbalized uncertainty}~\citep{tian2023just}: directly prompting the model to report its confidence, e.g., “I am 80\% confident that this answer is correct.” We refer to this approach as \emph{vanilla} uncertainty elicitation. Under well-controlled settings, vanilla performs reasonably well; 
however, several failures have been reported in practically important scenarios.

The first challenge arises in ambiguous question-answering scenarios, as illustrated in Figure~\ref{fig:concept}(a), where prompt underspecification admits multiple answers that may be simultaneously valid under different interpretations~\citep{min2020ambigqa, yang2025maqa}. In such cases, the degree of ambiguity should be faithfully reflected in the model’s verbalised uncertainty estimates. However, vanilla confidence measures and existing approaches often fail to reliably differentiate these situations; see Figure~\ref{fig:concept}(b).
The second challenge arises in in-context learning (ICL; \citet{brown2020language}), where task-specific examples are incorporated into the prompt without parameter updates. ICL is often interpreted as a form of implicit meta-learning~\citep{von2023transformers, wu2025why}, where the model infers the underlying task from the provided examples. As predictive performance often improves with additional examples, ICL is frequently described through a Bayesian epistemic uncertainty reduction~\citep{xie2022an, wakayama2025context}. Figure~\ref{fig:concept}(c) illustrates its failure mode: as more in-context examples are provided, the prediction error decreases, yet the uncertainty remains high and flat. This misalignment echoes \citet{falck2024is}, who argue from a martingale viewpoint that ICL is not strictly Bayesian.
The third challenge arises in self-reflection settings, where an LLM is prompted to select an answer from candidate options and subsequently reflect on its choice. Under a Bayesian-rationality, action selection should follow the maximization of expected utility as prescribed by Bayesian decision theory~\citep{harsanyi1978bayesian}. However, the utilities implicitly induced by elicited uncertainty scores often fail to account for the model’s observed decisions; see Figure~\ref{fig:concept}(d). This is consistent with \citet{liu2025large, yamin2026llms}, who rejects Bayesian rationality of LLMs through conditional independence testing.

\begin{figure}[t]
    \centering
    \includegraphics[width=1\linewidth]{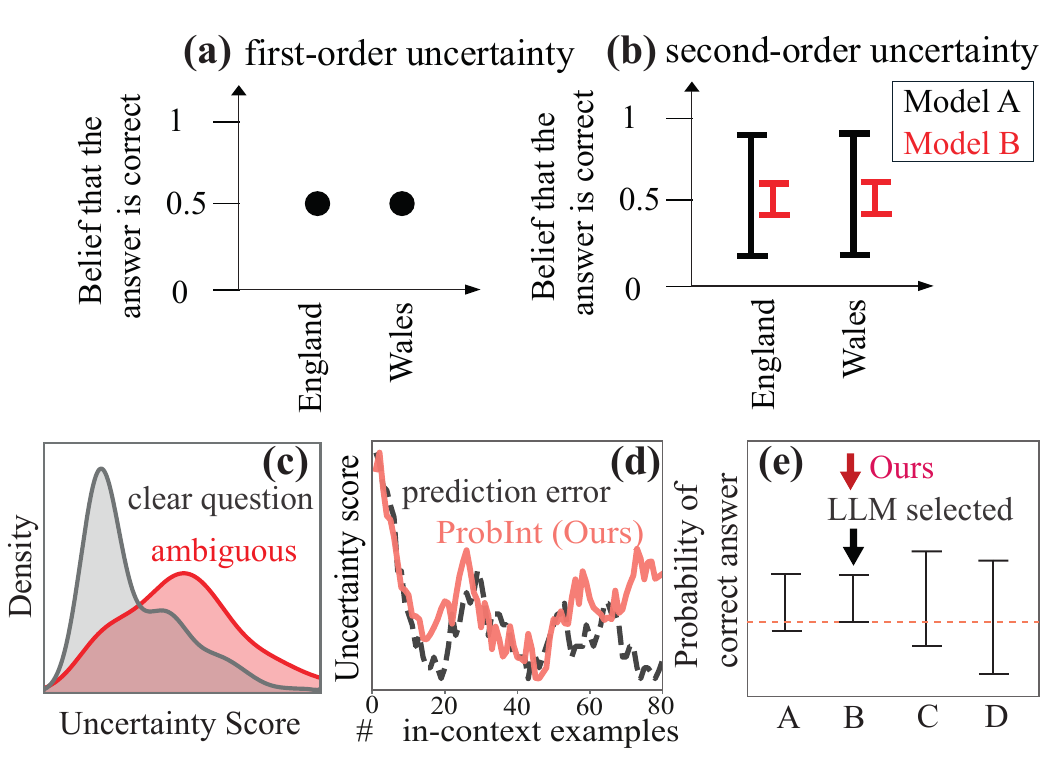}
    \caption{Our imprecise probabilities–based approach.
    (a) Classical precise probability provides point estimates.
    (b) Imprecise probabilities instead represents uncertainty as intervals.
    (c) This enables more reasonable elicitation of question ambiguity.
    (d) It more closely tracks predictive error.
    (e) It aligns the LLM’s answer selection with the selection implied by its imprecise probability estimates.}
    \label{fig:ip}
\end{figure}

These failures may stem not from an LLM’s inability to express uncertainty, but from the \emph{representation} we impose. Prior work implicitly assumes that uncertainty can be fully captured by a single, precise probability that rationally aggregates all sources into one value—and that such a score can be faithfully elicited. Instead, we ask: what if we allow LLMs to express uncertainty \emph{about} uncertainty—introducing imprecision into the uncertainty representation itself and building our foundation upon it? This is precisely the core of \emph{Imprecise Probabilities} (IP)~\citep{walley1991imprecise,augustin2014introduction}, a classical framework that enables the representation of higher-order uncertainty.

IP provides a principled foundation for understanding verbalized uncertainty in LLMs:
As illustrated in Figure~\ref{fig:ip}(a), conventional uncertainty scores reflect \emph{first-order uncertainty}: they provide point estimates that capture variability over outcomes (often associated with aleatoric uncertainty). In contrast, IP represents \emph{second-order uncertainty} through probability intervals, where any value within the interval is admissible. The interval widths quantify imprecision in the uncertainty estimate—uncertainty about uncertainty—commonly interpreted as epistemic uncertainty\footnote{As the terminology surrounding aleatoric/epistemic for LLMs remains under active debate, we adopt the terms first-/second-order uncertainty throughout for consistency~\citep{kirchhof2025position}.}. Through this lens, question ambiguity can be viewed as irreducible first-order uncertainty, whereas ICL reduces second-order uncertainty by providing additional examples that the model can use to improve prediction. Self-reflection can then be interpreted as decision-making under IP, typically adopting the argmax of the lower probability, i.e., the maximin rule~\citep{robbins1951asymptotically}. By leveraging the corresponding IP tools, the resulting uncertainty scores become substantially more coherent across tasks; see Figures~\ref{fig:ip}(c)–(e). 

While conceptually appealing, the adoption of IP frameworks for LLMs is far from straightforward. We present the first concrete instantiation of an IP-based approach for verbalized uncertainty in LLMs, introducing general-purpose prompting strategies and post-processing procedures designed to extract higher-order uncertainty from model outputs. Empirically, we show that the proposed framework not only yields a more coherent and rich uncertainty representations but also improves performance over existing methods while simultaneously incurring low API costs.

%% file: sections/02_method.tex
\section{Background}
We first review the background on LLM uncertainty quantification setting and imprecise probability, and then introduce our method for eliciting IPs from LLMs.

\subsection{LLM uncertainty quantification}
\textbf{Shared dataset structure.}
Assume the question-answer pair $(x,\mathcal{Y}^\star)$, where $x$ is a question prompt and $\mathcal{Y}^\star \subseteq \mathcal{Y}$ is the set of ground-truth correct answers, and $\mathcal{Y}$ is candidate answers. We use $y\in\mathcal{Y}$ to denote a candidate answer. 

\textbf{Tasks.}
Given question $x$, the task is to estimate the ground-truth answer $y^\star \in \mathcal{Y}^\star$. When the candidate set $\mathcal{Y}$ is not provided in the prompt, we refer to this as open-ended. 

\textbf{Language model.}
Let a pretrained LLM serve two roles:(i) \emph{Predictor}: $\hat{y} = \argmax_{y \in \mathcal{Y}} \hat{p}(y \mid x)$, producing a single answer estimate via the predictive distribution $\hat{p}(y \mid x)$; (ii) \emph{Generator}: $\hat{\mathcal{Y}} \sim \hat{p}(y \mid x)$, generating multiple candidate answers conditioned on the question $x$ when $\mathcal{Y}$ is not provided (prompt in Appendix~\ref{apd:approx_Yhat}).

\textbf{First-/second-order uncertainties.}
We define first-order uncertainty as intrinsic randomness arising from the question that the model cannot reduce. This system-level randomness occurs in cases such as $|\mathcal{Y}^\star| > 1$. Prior works~\citep{min2020ambigqa,yang2025maqa} consider the case $|\mathcal{Y}^\star| > 1$. All remaining uncertainty is attributed to second-order uncertainty. Equivalently, if $x$ contains sufficient information to uniquely determine the correct answer (i.e., $|\mathcal{Y}^\star|=1$), then first-order uncertainty is absent.

\subsection{Imprecise probabilities}\label{sec:bg_ip}
\textbf{Verbalized uncertainty.}
Our IP approaches elicit the model's beliefs by prompting the LLM to verbalize numerical uncertainty judgments (e.g. ~\citet{tian2023just}), rather than estimating uncertainty via repeated sampling from the model's outputs~\citep{farquhar2024detecting}.

\textbf{Compared to Bayesian.}
The Bayesian approach requires a prior distribution over the parameter space, and the distinction between aleatoric and epistemic uncertainty relies on an explicit likelihood specification. In contrast, IP treats any single-valued uncertainty score as first-order; a perfect estimator would coincide with pure aleatoric uncertainty. In practice, however, such perfection is unlikely, and most systems therefore exhibit residual second-order uncertainty, naturally represented within the IP framework.

\textbf{Ignorance vs indifference.}
Figure \ref{fig:ip}(b) contrasts two models, A and B. Model A exhibits wide probability intervals, reflecting ignorance—a lack of knowledge about which outcomes are plausible. Model B, in contrast, produces narrower intervals, indicating greater precision while still allowing for a degree of imprecision. By comparison, the first-order uncertainty representation in Figure \ref{fig:ip}(a) can only suggest that England and Wales are equally likely, without conveying any information about the underlying source of uncertainty. In particular, it cannot distinguish whether the model assigns equal probabilities due to genuine indifference (both outcomes considered equally plausible) or due to ignorance (insufficient information to prefer one over the other). Imprecise probability representations explicitly capture this distinction, separating ignorance from indifference.

\textbf{Probability intervals.}
IP provides rich representations to construct the probability interval~\citep{keynes1921atreatise}. The simplest approach is to directly prompt the model to report an interval $\big[\underline{p}(y), \overline{p}(y)\big]$ for each candidate answer $y$, where $\underline{p}(y)$ and $\overline{p}(y)$ denote the lower and upper probabilities. Intuitively, $\underline{p}(y)$ captures the probability that is \emph{certainly} justified by the evidence, while $\overline{p}(y)$ captures what is \emph{possibly} defensible \citep{augustin2014introduction}. Requesting such intervals tests whether the model can recognize and articulate the bounds of its belief; values between these bounds correspond to its personal ``medial'' probability for the event \citep{smith1961consistency}.

\textbf{Credal sets.}
Another approach is to elicit uncertainty from a \emph{group} of models, interpreting disagreement among them as imprecision. Each model is asked to report its \emph{first-order} uncertainty for a given answer $y$. The collection of these reported probabilities forms a credal set \citep{levi1980enterprise}, and the minimum and maximum values across the set define the corresponding probability interval.

\textbf{Possibility.}
The last alternative is via exclusion: it evaluates whether the model can confidently rule out a candidate answer $y$ by assessing the \emph{possibility} of alternative answers, which is called \emph{a possibility function} $\pi(y) \in [0,1]$~\citep{dubois1985theorie}. Possibility functions, unlike probabilities, are non-additive and are normalized only by requiring that at least one output is fully plausible (i.e. $\smash{\sup_y\pi(y)=1}$). This supports relative plausibility comparisons without allocating a fixed mass across candidates—making elicitation less sensitive to systematic miscalibration and overconfidence in model self-reports~\citep{wang2025dinco,xiong2024can}—and lets us elicit $\pi$ for “none of the above” without changing the plausibility assigned to other outputs.

\section{Uncertainty Elicitation via Imprecise Probabilities}\label{sec:ip}

We now introduce our methods to elicit higher-order uncertainties from LLMs via prompting and post-processing strategies. We introduce the specific prompting techniques for both first- and second-order uncertainties.

\subsection{First-order Uncertainty}
Under IP, any point-valued score (including the vanilla) is considered first-order; however, it can be refined to better satisfy the probability axioms (non-negativity, additivity, normalization)~\citep{kolmogorov1933}. We draw on Bruno de Finetti’s classical interpretation of probability as coherent betting behavior \citep{definetti1937foresight}. Under this view, a probability corresponds to the fair price at which an agent is willing to buy or sell a gamble. Rational betting prices must satisfy the probability axioms, as violations would expose the agent to a sure loss (i.e., a Dutch book).
This betting-based interpretation is directly implementable through Prompt~\ref{prompt:definetti} and Alg.~\ref{alg:definetti}. 
The probability-axiom verifier algorithmically enforces compliance with the axioms. Given elicited betting prices $p_k \in \hat{p}$ over mutually exclusive and exhaustive candidate answers, coherence reduces to verifying: (i) non-negativity, $p_k \geq 0$, and (ii) normalization, $\sum_{k=1}^{\lvert \mathcal{Y} \rvert} p_k = 1$. Under this assumption, additivity follows automatically.

\subsection{Second-order Uncertainty}

\begin{figure}[t]
\centering
\begin{promptbox}[DeFinetti]{prompt:definetti}
    Assign a buy price (between \$0.00 and \$1.00) for each answer representing the maximum amount you would pay for a bet on that answer being correct.
    If an answer is correct, the bet pays \$1.00; if incorrect, it pays \$0.00, and the price paid is lost.
    Assign prices that maximize expected profit, taking into account how each answer might be correct or incorrect under reasonable alternative interpretations of the question (e.g., unclear entities, ambiguous events, or uncertainty about required answer format or type), and how multiple answer options can be equally correct.
    The prices must sum to exactly \$1.00 across all answers.
\end{promptbox}
\end{figure}

\begin{algorithm}[t!]
\caption{DeFinetti}
\label{alg:definetti}
\begin{algorithmic}[1]
\STATE \textbf{Input:} input $x$, candidate answers $\mathcal{Y}$, verifier $C_\text{axiom}$
\STATE \textbf{Init:} $\hat{p}(y \mid x) \gets 0$ for all $y \in \mathcal{Y}$.
\WHILE{$C_\text{axiom}[\hat{p}(\mathcal{Y} \mid x)]$ is \texttt{False}}
    \STATE $\hat{p}(\mathcal{Y} \mid x) \gets \text{DeFinettiBet}(x, \mathcal{Y})$.
\ENDWHILE
\STATE \textbf{return:} Entropy $H(\hat{p})$
\end{algorithmic}
\end{algorithm}
Next, we elicit second-order uncertainty via IP representations.
As shown in Alg.~\ref{alg:ip}, the overall framework remains unchanged; the key differences are: (i) replacing point estimates $\hat{p}$ with probability intervals $[\underline{p}, \overline{p}]$, and (ii) replacing entropy with the maximum mean imprecision (MMI) metric \citep{chau2025integral}. We first describe how to elicit probability intervals, and then introduce the MMI metric.

\textbf{Representations.}
As discussed in §\ref{sec:bg_ip}, IP provides three principal representations of uncertainty: probability intervals, credal sets, and possibility measures. A natural question is which representation to adopt. In general, these capture different aspects of uncertainty, and no single representation is universally superior.
We adopt the following perspective:
(i) probability intervals serve as the most basic and widely applicable representation;
(ii) credal sets are particularly suitable when using an ensemble of LLMs, and otherwise remain optional; and
(iii) possibility measures are useful when the candidate set $\mathcal{Y}$ may be incomplete (e.g., in open-ended Q\&A settings), and are otherwise optional.

\textbf{Probability interval.}
Following Alg.~\ref{alg:ip}, probability intervals can be obtained by replacing the IP-Prompt in Line~\ref{alg_line:ip-prompt} with Prompt~\ref{prompt:interval}. The verifier $C_{\text{verifier}}$ checks if (i) the lower probabilities satisfy $\sum_y \underline{p}(y) \leq 1$, and (ii) the upper probabilities satisfy $\sum_y \overline{p}(y) \geq 1$ \footnote{Since the upper-bound MMI depends only on $\underline{p}$, we verify constraint (i) only.}. We name this as \textsc{ProbInt}.

\begin{figure}[t]
\centering
\begin{promptbox}[ProbInt]{prompt:interval}
Provide a lower and upper probability (each between 0.0 and 1.0) indicating how likely the answer is correct. Interpret the probabilities as follows:\\
• \textbf{Lower Probability}: the smallest probability you consider plausible that the answer is correct.\\
• \textbf{Upper Probability}: the largest probability you consider defensible that the answer is correct.\\
The sum of all lower probabilities across all answers must not exceed 1.0.
\end{promptbox}
\end{figure}

\textbf{Other representations.}
For credal sets, we form an ensemble of LLMs—either distinct models or multiple seeded runs of the same model (see Prompt~\ref{prompt:credal}). For possibility functions, see Prompt~\ref{prompt:possibility}. We name each method as \textsc{Credal} and \textsc{Pos}.

\subsection{Maximum Mean Imprecision}\label{sec:mmi}
\begin{algorithm}[t!]
\caption{Imprecise probability}
\label{alg:ip}
\begin{algorithmic}[1]
\STATE \textbf{Input:} input $x$, candidate answers $\mathcal{Y}$, verifier $C_\text{verifier}$
\STATE \textbf{Init:} $\hat{p}(y \mid x) \gets 0$ for all $y \in \mathcal{Y}$.
\WHILE{$C_\text{verifier}[\hat{p}(\mathcal{Y} \mid x)]$ is \texttt{False}}
    \STATE $\underline{p}(\mathcal{Y} \mid x), \,\, \overline{p}(\mathcal{Y} \mid x) \gets \text{IP-Prompt}(x, \mathcal{Y})$. \label{alg_line:ip-prompt}
\ENDWHILE
\STATE \textbf{return:} $\text{MMI}(\underline{p}, \overline{p})$
\end{algorithmic}
\end{algorithm}
We compute second-order uncertainty as a scalar metric from IP representations using MMI~\citep{chau2025integral} with total variation. Since exact computation scales exponentially with $|\mathcal{Y}|$ (Appendix~\ref{apd:mmi}), we employ suitable approximations.

\textbf{Intuition.}
The most straightforward uncertainty score derived from an IP representation is the interval width:
\begin{equation}\label{eq:mmi_binary}
    \text{MMI} =\overline{p}(y)-\underline{p}(y).
\end{equation}
This expression is exact for a single answer $y$. However, when $\lvert \mathcal{Y} \rvert > 1$, aggregating multiple intervals into a single scalar measure requires the formal MMI definition, which accounts for interactions across candidates.

\textbf{Upper bound.}
\citep{chau2025integral} provides an easily computable upper bound:
\begin{equation} \label{eq:mmi_upper}
\text{MMI} \leq 1-\sum_{y\in\mathcal{Y}}\underline{p}(y).
\end{equation}
We use this upper bound as a tractable approximation of the exact MMI. We include additional details in Appendix~\ref{apd:mmi}.

\subsection{Related works}

\textbf{Verbalized uncertainty.} Vanilla verbalized uncertainty is known to be overconfident~\citep{tian2023just,xiong2024can}. Subsequent work has focused on improving calibration, including alternative prompting and elicitation styles~\citep{tian2023just,xiong2024can}, calibration-oriented fine-tuning~\citep{li2025conftuner,lin2022teaching,xu2024sayself,stengeleskin2024lacie}, post-hoc adjustments such as normalization~\citep{wang2025dinco}, and language-based uncertainty expressions~\citep{kirchhof2025self}. 
However, prior work typically assumes no question ambiguity, whereas our framework explicitly separates first-/second-order uncertainty.

\textbf{Uncertainty disentanglement.} Disentangling first-/second-order uncertainties has increasingly been adapted to LLMs, but all prior work focused on sampling-based approach or assumption to the access to internal parameters: (i) Information-theoretic decompositions~\citep{lakshminarayanan2017deepensemble} have been reframed using generated clarification prompts, measuring how much clarifications reduce output uncertainty via mutual information~\citep{hou2024decomposing,walha2025finegrained,xia2025survey}. (ii) Perturbations: varying temperature and prompts, measure its change as uncertainty score~\citep{gao2024spuq}, and (iii) via linear probes on internal activations~\citep{ahdritz2024distinguishing}. We instead target disentanglement via \emph{verbalized} uncertainty, which to our knowledge has not been done.

\textbf{Human experts elicitation}
IP methods originate from frameworks developed to model and elicit human judgments. Similar elicitation-based methodologies are now increasingly adopted in LLM research, including preference learning \citep{rafailov2023direct, fujisawa2025scalable}, rubric-based evaluation frameworks \citep{huang2025reinforcement}, item response theory \citep{polo2024tinybenchmarks, mencattini2025merge}, and social choice-theoretic approaches \citep{muandet2022impossibility, adachi2025bayesian}.

%% file: sections/03_synthetic_exp.tex
\section{Synthetic Experiment}\label{sec:synthetic}
\begin{figure}[t]
    \centering
    \includegraphics[width=0.85\linewidth]{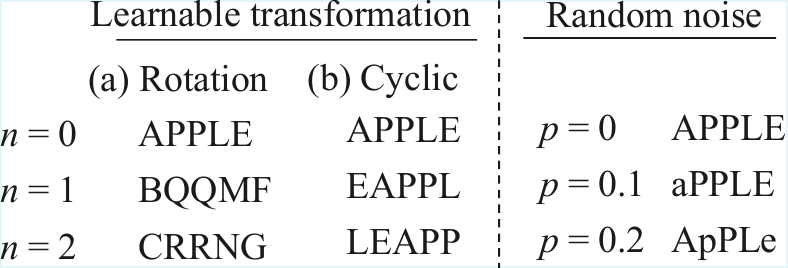}
    \caption{Learnable vs. noisy transforms.}
    \label{fig:syn}
\end{figure}
Our goal is to elicit an LLM’s uncertainty through conversational prompts. However, uncertainty estimation is inherently unsupervised and thus difficult to evaluate objectively. To address this, we first construct a synthetic dataset with an explicitly controlled data-generating process in this section. We then apply our approach to real-world datasets, following standard practice (§\ref{sec:real-world}).

\begin{figure}[t]
  \centering
  \includegraphics[width=0.95\linewidth]{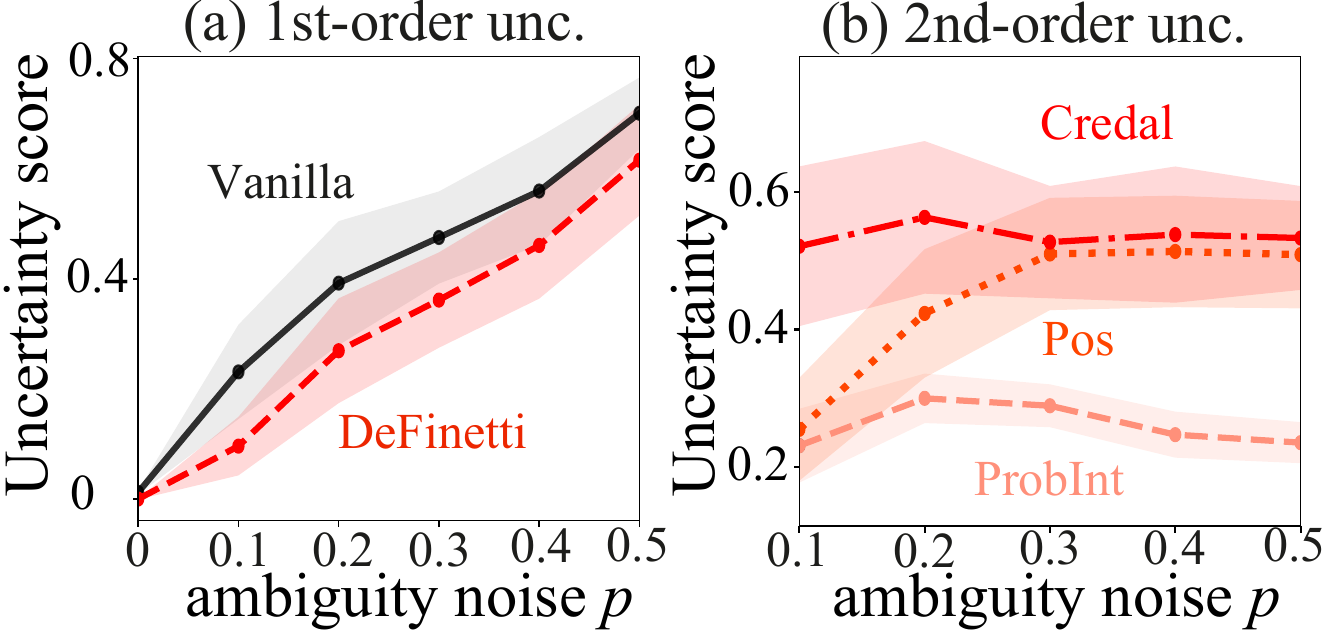}
  \caption{(a) For first-order uncertainty estimation, both vanilla and De Finetti capture the underlying ambiguity noise $p$, but (b) for second-order, our methods stay flat, supporting the disentanglement of uncertainty source.}
  \label{fig:au_noise}
\end{figure}
\textbf{Data generation.}
Following \citet{zhao2025chain}, we construct synthetic sequence-transformation tasks (Figure~\ref{fig:syn}), including (a) rotation and (b) cyclic shift, each parameterized by $n$. We additionally inject noise by randomly lowercasing each output letter with probability $p$.

\textbf{Task.}
Given an input string $x$, the model predicts its transformed output $y$ (e.g., APPLE $\rightarrow$ BQQMF). The LLM receives $m$ example pairs ${(x_j, y_j)}_{j=1}^m$ generated from a consistent transformation rule and must infer this rule to predict $\hat{y}_{m+1}$ for a new input $x_{m+1}$. We adopt ICL, where examples are provided via prompts without parameter updates.

\textbf{Metric.}
We evaluate prediction using a \emph{permissive} string match: predictions are first capitalized, then compared exactly with the ground-truth all-capitalized $y^\star_{m+1}$. Under this evaluation, capitalization differences are ignored, and the model is evaluated on whether it recovers the underlying rotation and cyclic-shift rules. Equivalently, all capitalization variants of the correct answer are treated as valid: any $\hat{y}_{m+1}\in\mathcal{Y}^\star_{m+1}$ is judged correct. The parameter $p$ controls the amount of capitalization-induced ambiguity, which we refer to as ambiguity noise.

\textbf{Controlling first-/second-order uncertainty.}
We treat the ambiguity noise $p$ as the first-order because $\lvert \mathcal{Y}^\star_{m+1} \rvert > 1$, and the number of in-context examples $m$ controls the second-order as $m$ augments knowledge, while first-order noise remains constant \citep{ling2024uncertainty, wang2025uncertainty}.

\textbf{Baselines.}
We compare against \textsc{Vanilla}~\citep{tian2023just}, which directly elicits a confidence probability on a predicted answer $\hat{y}$. For evaluation, we convert confidence to an uncertainty score using $1-\mathrm{conf}$, a monotone transformation that preserves the ranking.

\textbf{Base setup.}
We fix the transformation to rotation ($n=13$), followed by cyclic shift ($n=1$), and random lowercasing with probability $p$. For each $m$, we generate five independent ICL example sets and evaluate each with five random seeds (25 runs total), accounting for variability in both ICL examples and stochastic model outputs. Unless otherwise specified, we use \textsc{gpt-5-mini}~\citep{achiam2023gpt4tr}.

\subsection{First-order noise}
We evaluate uncertainty disentanglement under ambiguity noise levels $p \in \{0,0.1,0.2,0.3,0.4,0.5\}$. To isolate ambiguity effects, we fix the number of ICL examples at $m=80$. Figure~\ref{fig:au_noise} reports the behavior of each uncertainty score for (a) first- and (b) second-order.
As expected, both \textsc{Vanilla} and \textsc{DeFinetti-AU} increase approximately linearly with $p$, while our IP methods remain largely invariant. \textsc{Pos} shows some sensitivity to $p$ because it relies on a scalar possibility value, unlike interval-based approaches such as \textsc{Credal} and \textsc{ProbInt}. 
Nevertheless, at higher $p$—when alternative answers become equally more plausible—\textsc{Pos} stabilizes and becomes 
less sensitive to further increases in $p$.

\subsection{Second-order denoise}\label{sec:icl_eu}
\begin{figure}[t]
  \centering
  \includegraphics[width=0.85\linewidth]{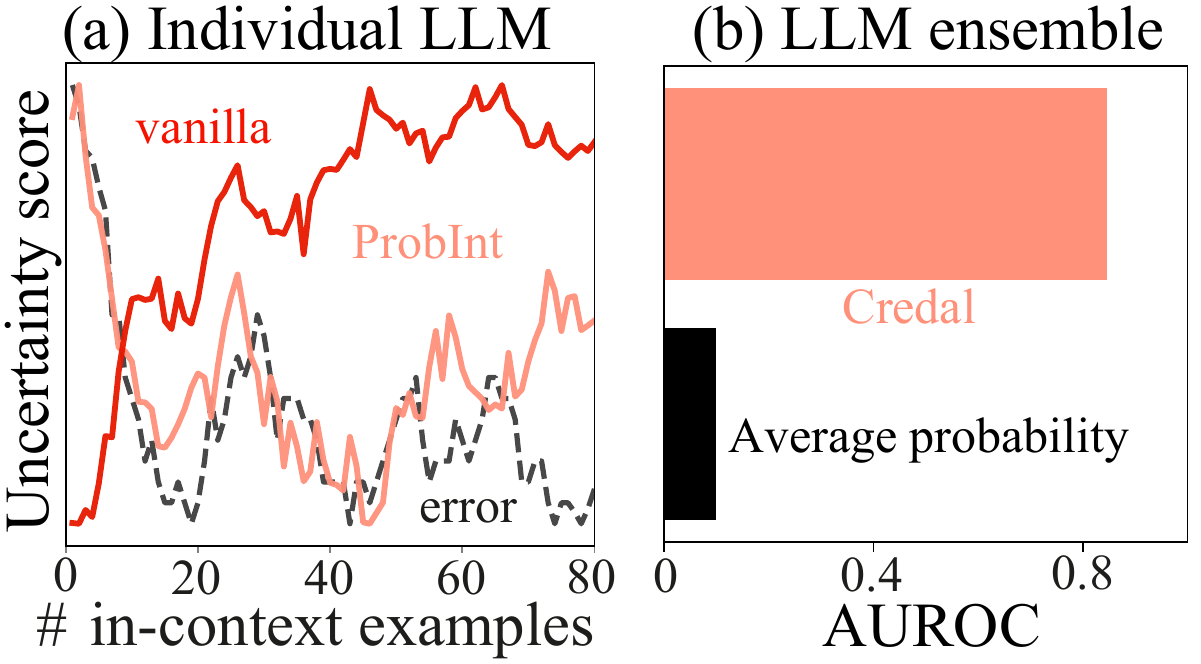}
  \caption{(a) For individual EU elicitation, \textsc{vanilla} incorrectly increases uncertainty, whereas \textsc{ProbInt-EU} tracks prediction error. (b) For group EU elicitation, our \textsc{Credal-EU} substantially improves AUROC.}
  \label{fig:au_fixed_eu}
\end{figure}
\input{results/eu_no_au_gpt5}

Next, we fix $p=0.25$ and vary the number of ICL examples $m$. Since increasing $m$ reduces second-order uncertainty, a well-calibrated score should decrease accordingly. As shown in Figure~\ref{fig:au_fixed_eu}, \textsc{ProbInt} closely tracks prediction error, whereas \textsc{Vanilla} remains overly uncertain and fails to reflect the error reduction. Results for the credal-set and possibility-based methods are reported in the Appendix~\ref{apd:icl}; both exhibit decreasing trends.

\subsection{Group uncertainty elicitation} \label{sec:icl_ensemble}
\begin{figure}
    \centering
    \includegraphics[width=1\linewidth]{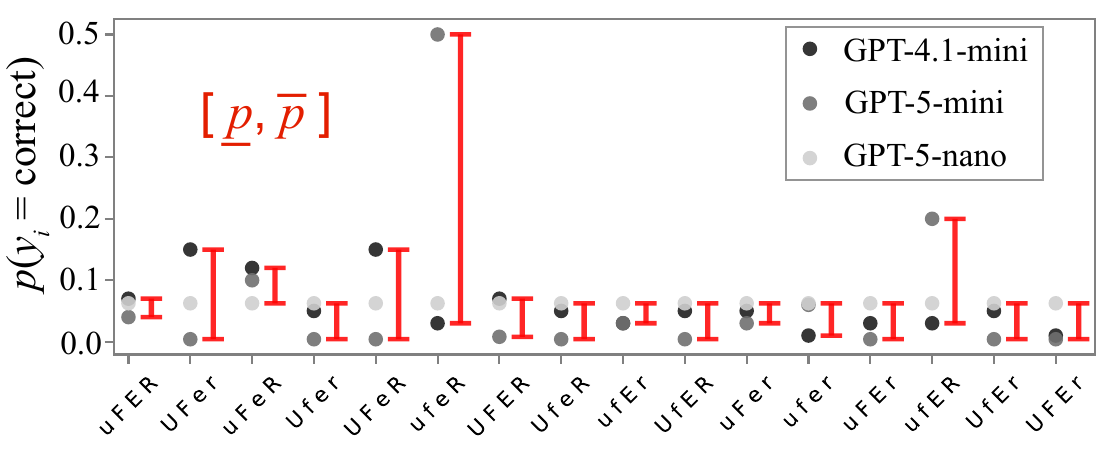}
    \caption{Credal-set from LLM ensemble: model disagreement induces lower/upper probability bounds for each output, which are used to compute MMI.}
    \label{fig:credal_prob_int}
\end{figure}

Next, we consider group uncertainty elicitation for an LLM ensemble. As LLMs are increasingly deployed in multi-agent systems (e.g., \citep{lange2026shinkaevolve}), collective uncertainty becomes crucial. The credal-set view provides a natural framework, as it captures cross-model disagreement.
In Figure~\ref{fig:credal_prob_int}, the $x$-axis enumerates correct answers under the transformation that differ only by the lowercase noise $p$. The probability of each answer under the noise varies across candidates $y_i$. Let $\ell$ denote the number of lowercase characters in $y_i$; then $p(y_i) = p^\ell \cdot (1-p)^{1-\ell}$. With $p=0.25$, the ideal probabilities are $p(y_i) \in \{0.316, 0.105, 0.035, 0.012, 0.004\}$ for $\ell \in \{0,1,2,3,4\}$. In practice, these values depend on the ICL examples and may deviate from the ideal case.

We apply the \textsc{Credal} method to elicit group uncertainty. 
Unlike the single-LLM setting, where $\hat{y}$ can be obtained simply by asking the model which option to choose, the group setting requires an additional decision rule to aggregate votes across multiple LLMs. To avoid this added complexity, we instead compute AUROC against $p(y_i = \text{correct})$ for each answer and use the exact MMI (Eq.~\ref{eq:mmi_binary}) at the answer level.
To compute MMI, we first elicit each LLM to report its \emph{first-order} probabilities $p(y_i = \text{correct})$ (dots in Fig. \ref{fig:credal_prob_int}). The IP intervals (red bars) are obtained by taking the pointwise minimum/maximum probabilities across models. 
We compare \textsc{Credal} with the standard aggregation method used in LLM ensembles—utilitarian aggregation—defined as, 
$\mathbb{E}_{\mathcal{M}_i}[p(y = \text{correct} \mid \mathcal{M}_i)]$. Taking the argmax over this quantity recovers popular strategies such as majority voting or best-of-$N$ (e.g., \citep{wang2022self, si2023prompting}). In our evaluation, however, we use the averaged score directly and compare it with our MMI score using AUROC.
As shown in Figure~\ref{fig:au_fixed_eu}(b), simple average provide limited predictive power, whereas \textsc{Credal} yields substantial improvement.

%% file: results/eu_no_au_gpt5.tex
\begin{table*}
    \centering
    \caption{AUROC for correctness detection without ambiguity. Best result bolded in \textcolor{blue}{\textbf{blue}}, second best in \textcolor{orange}{\textbf{orange}}.}\label{tab:aleatoric_uncertainty_gpt5_question}
    \resizebox{\textwidth}{!}{%
    \begin{tabular}{lccccccc}
        \toprule 
        &\multicolumn{3}{c}{GPT-5} & \multicolumn{3}{c}{gemini-2.5-pro} & \\
        \cmidrule(r){2-4} 
        \cmidrule(l){5-7}
        \bfseries Method & \bfseries MMLU-Pro & \bfseries Non-MAQA & \bfseries Non-AmbigQA & \bfseries MMLU-Pro & \bfseries Non-MAQA & \bfseries Non-AmbigQA & \bfseries Avg. Rank\\
        \midrule 
        Is true prob. & 0.6850 {\scriptsize$\pm$ 0.0238} & 0.5835 {\scriptsize$\pm$ 0.0041} & 0.5584 {\scriptsize$\pm$ 0.0157} & 0.6583 {\scriptsize$\pm$ 0.0197} & 0.6296 {\scriptsize$\pm$ 0.0068} & 0.5897 {\scriptsize$\pm$ 0.0057} & 8.67\\
        Label prob. & 0.7649 {\scriptsize$\pm$ 0.0220} & 0.6430 {\scriptsize$\pm$ 0.0062} & 0.5876 {\scriptsize$\pm$ 0.0176} & 0.6377 {\scriptsize$\pm$ 0.0153} & 0.6483 {\scriptsize$\pm$ 0.0126} & 0.5582 {\scriptsize$\pm$ 0.0319} & 7.83\\
        MI-Clarifications & 0.7790 {\scriptsize$\pm$ 0.0114} & 0.6528 {\scriptsize$\pm$ 0.0080} & 0.5840 {\scriptsize$\pm$ 0.0107} & 0.6348 {\scriptsize$\pm$ 0.0144} & 0.6546 {\scriptsize$\pm$ 0.0104} & 0.5497 {\scriptsize$\pm$ 0.0267} & 7.83\\
        DiNCO & 0.7087 {\scriptsize$\pm$ 0.0214} & 0.5642 {\scriptsize$\pm$ 0.0066} & 0.5219 {\scriptsize$\pm$ 0.0035} & 0.7133 {\scriptsize$\pm$ 0.0167} & 0.5749 {\scriptsize$\pm$ 0.0224} & 0.5275 {\scriptsize$\pm$ 0.0190} & 9.17 \\
        Vanilla & 0.8587 {\scriptsize$\pm$ 0.0192} & 0.7577 {\scriptsize$\pm$ 0.0194} & \textcolor{orange}{\textbf{0.7756 {\scriptsize$\pm$ 0.0138}}} & 0.7326 {\scriptsize$\pm$ 0.0221} & 0.6889 {\scriptsize$\pm$ 0.0263} & 0.6688 {\scriptsize$\pm$ 0.0209} & 4.17 \\
        Top-4 & 0.8725 {\scriptsize$\pm$ 0.0143} & 0.7668 {\scriptsize$\pm$ 0.0089} & 0.7666 {\scriptsize$\pm$ 0.0135} & 0.7652 {\scriptsize$\pm$ 0.0226} & \textcolor{orange}{\textbf{0.7027 {\scriptsize$\pm$ 0.0294}}} & 0.6698 {\scriptsize$\pm$ 0.0087} & 3.33\\
        CoT & 0.8548 {\scriptsize$\pm$ 0.0124} & 0.7695 {\scriptsize$\pm$ 0.0078} & 0.7724 {\scriptsize$\pm$ 0.0101} & 0.6826 {\scriptsize$\pm$ 0.0409} & 0.6431 {\scriptsize$\pm$ 0.0140} & 0.6094 {\scriptsize$\pm$ 0.0152} & 5.33 \\
        \\[-10pt]
        \textbf{ProbInt (ours)} & 0.8617 {\scriptsize$\pm$ 0.0082} & \textcolor{orange}{\textbf{0.7709 {\scriptsize$\pm$ 0.0058}}} & 0.7713 {\scriptsize$\pm$ 0.0044} & \textcolor{orange}{\textbf{0.7857 {\scriptsize$\pm$ 0.0285}}} & \textcolor{blue}{\textbf{0.7303 {\scriptsize$\pm$ 0.0092}}} & \textcolor{blue}{\textbf{0.7128} {\scriptsize$\pm$ 0.0187}} & \textcolor{blue}{\textbf{2.33}}\\
        \textbf{Credal (ours)} & \textcolor{blue}{\textbf{0.8798} {\scriptsize$\pm$ 0.0121}} & \textcolor{blue}{\textbf{0.7753} {\scriptsize$\pm$ 0.0067}} & \textcolor{blue}{\textbf{0.7932} {\scriptsize$\pm$ 0.0126}} & 0.7466 {\scriptsize$\pm$ 0.0201} & 0.6978 {\scriptsize$\pm$ 0.0199} & 0.5982 {\scriptsize$\pm$ 0.0469} & \textcolor{orange}{\textbf{2.67}}\\
        \textbf{Pos (ours)} & \textcolor{orange}{\textbf{0.8738 {\scriptsize$\pm$ 0.0216}}} & 0.7230 {\scriptsize$\pm$ 0.0179} & 0.7426 {\scriptsize$\pm$ 0.0228} & \textcolor{blue}{\textbf{0.7863} {\scriptsize$\pm$ 0.0163}} & 0.6806 {\scriptsize$\pm$ 0.0114} & \textcolor{orange}{\textbf{0.6780 {\scriptsize$\pm$ 0.0094}}} & 3.67\\
        \bottomrule 
    \end{tabular}
    }
\end{table*}

%% file: sections/04_realworld_exp.tex
\section{Real-world QA experiment}\label{sec:real-world}
We evaluate our methods on real-world QA benchmarks, following standard practice. 

\textbf{Datasets.}
We use two dataset types. (i) Open-ended QA: MAQA~\citep{yang2025maqa} and AmbigQA~\citep{min2020ambigqa}. Both contain a mix of ambiguous ($\lvert \mathcal{Y}^\star \rvert > 1$) and unambiguous items ($\lvert \mathcal{Y}^\star \rvert = 1$). (ii) Non-open-ended QA: MMLU-Pro~\citep{wang2024mmlu} (answer options $\mathcal{Y}$ is provided) but only contains unambiguous items  ($\lvert \mathcal{Y}^\star \rvert = 1$).

\textbf{Task.} Given question $x$, we prompt the LLM to predict a single answer $\hat{y}$. Correctness is determined by comparing $\hat{y}$ with a reference answer $y^\star$, defined as:
(i) non-ambiguous: $\mathcal{Y}^\star = \{y^\star\}$; or
(ii) ambiguous: a specific sample $y^\star \in \mathcal{Y}^\star$. Any $y \in \mathcal{Y}^\star$ but $y \neq y^\star$ is judged as incorrect.

\textbf{Metric.}
We report AUROC using each uncertainty score. (i) ambiguity: labels are $|\mathcal{Y}^\star|>1$ vs. $|\mathcal{Y}^\star| = 1$. (ii) correctness: labels are $\hat{y}=y^\star$ vs. $\hat{y}\neq y^\star$. AUROC is calculated by how well the uncertainty score separates the two classes, with higher uncertainty score corresponding to ambiguous examples for (i) and incorrect predictions for (ii).

\textbf{Baselines.}
The only directly comparable baseline is \textsc{MI Clarifications}~\citep{hou2024decomposing}, which decomposes total uncertainty (equivalent to semantic entropy~\citep{farquhar2024detecting}) into aleatoric (first-order) and epistemic (second-order) uncertainty via the mutual information between sampled outputs and generated clarifications. We restrict our comparison to black-box compatible baselines.

\textbf{Baselines for ambiguity.}
(i) \textsc{Semantic Entropy}~\citep{farquhar2024detecting}, entropy over semantically clustered samples;
(ii) \textsc{Ask4Conf-D}~\citep{hou2024decomposing}, directly eliciting the probability that a question is ambiguous; and
(iii) \textsc{MI Clarifications}~\citep{hou2024decomposing}.

\textbf{Baselines for correctness.}
We consider verbalized and sampling-based methods.
\emph{Verbalized:}
(i) \textsc{Vanilla}~\citep{tian2023just,xiong2024can};
(ii) \textsc{CoT}~\citep{xiong2024can}, eliciting a rationale to improve prediction and confidence; and
(iii) \textsc{Top-4}~\citep{tian2023just}, querying the top four answers but using only the top answer’s confidence.
\emph{Sampling-based:}
(i) \textsc{Is-True}~\citep{tian2023just,kadavath2022languagemodelsmostlyknow};
(ii) \textsc{Label Probability}, estimating an empirical answer distribution from samples;
(iii) \textsc{DiNCO}~\citep{wang2025dinco}, extending \textsc{Is-True} with distractors and normalization, and 
(iv) \textsc{MI Clarifications}~\citep{hou2024decomposing}.

\textbf{Base setup.}
For each dataset, we sample $N=200$ examples from the validation set and repeat experiments five times, reporting mean and standard deviation. Sampling-based baselines and \textsc{Credal} use five samples per query; runs with different seeds are treated as distinct beliefs.

\subsection{Ambiguity and correctness}
\begin{figure}
    \centering
    \includegraphics[width=\linewidth]{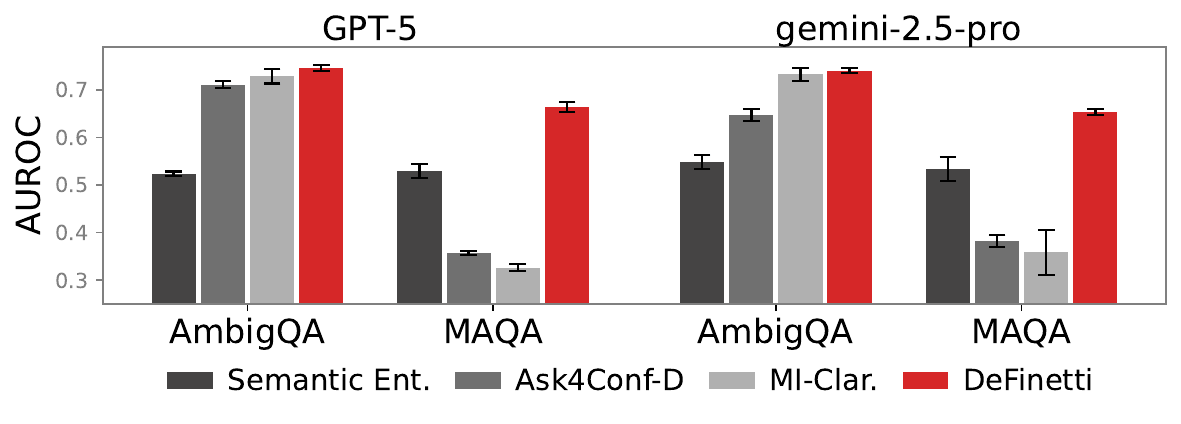}
    \caption{AUROC for ambiguity detection.}
    \label{fig:au_auroc}
\end{figure}

\textbf{Ambiguity detection.} 
As shown in Fig.~\ref{fig:au_auroc}, our \textsc{DeFinetti} achieves the highest AUROC for detecting ambiguity. Interestingly, directly eliciting an ambiguity probability (\textsc{Ask4Conf-D}) underperforms on MAQA, suggesting that a single global ambiguity judgment is insufficient without modeling the answer distribution.

\textbf{Correctness detection.}
We next evaluate correctness detection capability. We first select unambiguous subsets from MAQA and AmbigQA (i.e., $|\mathcal{Y}^\star|=1$), and we call them as Non-MAQA and Non-AmbigQA. Results are reported in Table~\ref{tab:aleatoric_uncertainty_gpt5_question}.
Overall, our methods perform best, although verbalization-based approaches also work reasonably well in this setting. This is consistent with prior findings: in the absence of ambiguity, uncertainty scores provably perform reliably~\citep{tomov2025illusion}.
Empirically, \textsc{ProbInt} is the most robust across datasets and models.

\begin{figure}
    \centering
    \includegraphics[width=\linewidth]{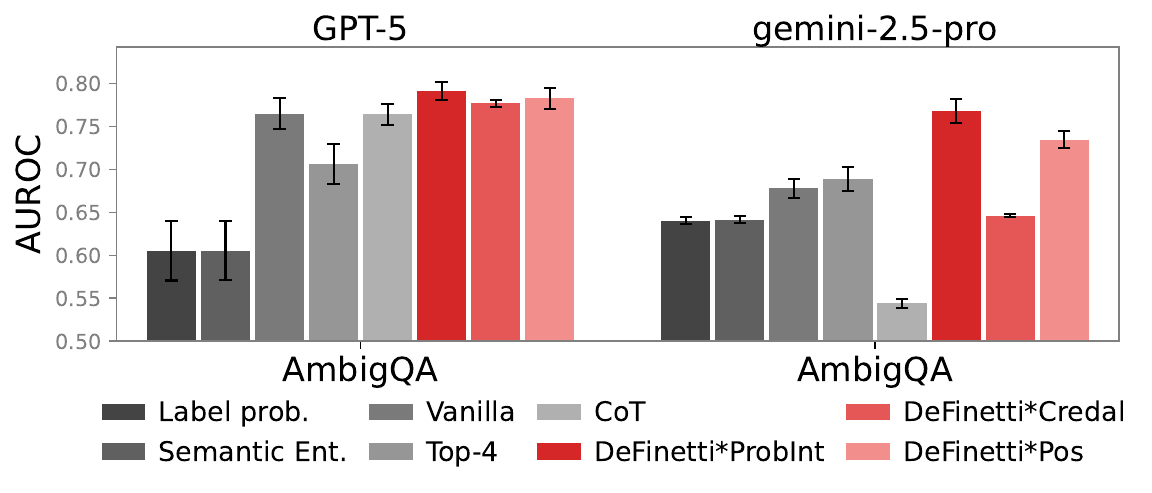}
    \caption{AUROC for correctness with ambiguity.}
    \label{fig:auroc_total}
\end{figure}

\textbf{Correctness detection under ambiguity.}
Under simultaneous ambiguity and incomplete knowledge, isolated scores may fail to capture total uncertainty. Since \textsc{DeFinetti} and IP scores are measured on different scales, they cannot be directly summed. We therefore combine them multiplicatively to achieve scale invariance. Figure~\ref{fig:auroc_total}\footnote{We take $y^\star$ to be the reference answer from Natural Questions~\citep{kwiatkowski2019natural}, included in AmbigQA.} shows that this product separates correct from incorrect predictions $\hat{y}$ better than the baselines under combined uncertainty.

\textbf{API cost.}
\begin{figure}[b!]
    \centering
    \includegraphics[width=0.90\linewidth]{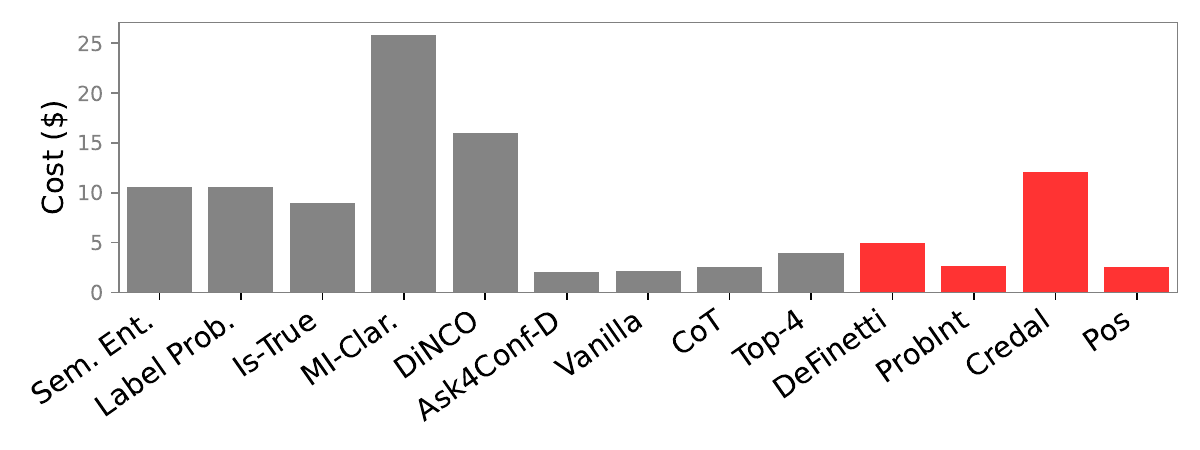}
    \caption{API cost on MMLU-Pro across all methods.}
    \label{fig:api_cost}
\end{figure}
We report API costs computed from public pricing for all methods (Fig.~\ref{fig:api_cost}). Compared to sampling-based baselines, our methods are generally more cost-efficient, except \textsc{Credal} which also uses sampling. \textsc{ProbInt} and \textsc{Pos} are comparable in cost to verbalized baselines. Compared to \textsc{MI Clarifications}—the only baseline that disentangles uncertainty—our methods cost less than half.

\begin{figure}[t]
    \centering
    \includegraphics[width=0.9\linewidth]{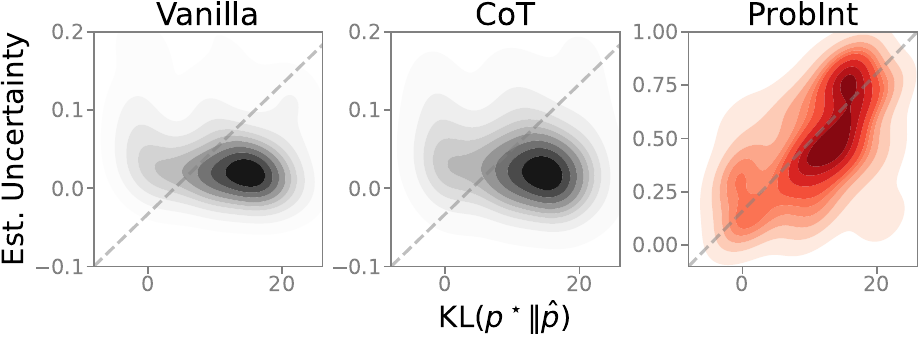}
    \caption{\textsc{ProbInt} more closely matches the KL metric.}
    \label{fig:maqa_eu_with_au}
\end{figure}
\begin{figure}[t]
    \centering
    \includegraphics[width=\linewidth]{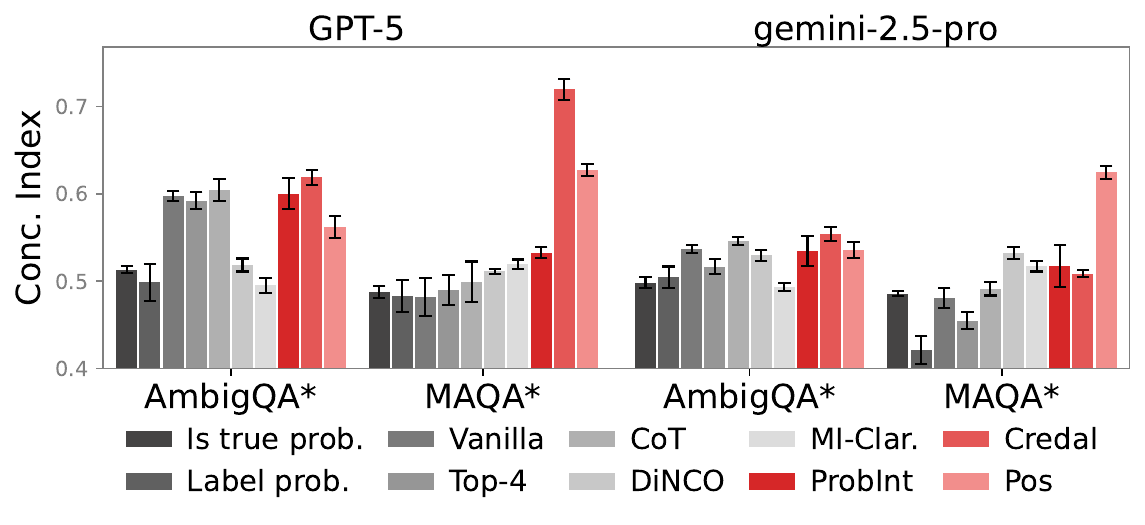}
    \caption{Concordance index ($\uparrow$ better) between uncertainty scores and the KL metric.}
    \label{fig:eu_with_au1}
\end{figure}

\textbf{Visualization.}
For ambiguity, we use the MAQA dataset. Figure~\ref{fig:ip}(c) shows that our \textsc{DeFinetti} estimate effectively separates ambiguous from clear questions, whereas the \textsc{Vanilla} score in Figure~\ref{fig:concept}(b) fails to do so.
For correctness, we adopt the Kullback–Leibler (KL) divergence metric proposed by \citet{tomov2025illusion}, who construct a reference answer distribution $p^\star$ from question–answer co-occurrence frequencies in a large corpus. They evaluate uncertainty (in the absence of ambiguity) via the KL divergence between the LLM’s predictive distribution $\hat{p}$ and $p^\star$ (see Appendix~\ref{apd:real}). Figure~\ref{fig:maqa_eu_with_au} shows that our \textsc{ProbInt} method exhibits the strongest correlation with this KL metric.
We further compute the concordance index, which measures rank-based correlation between two variables, again comparing the KL metric with uncertainty scores. As shown in Figure~\ref{fig:eu_with_au1}, our IP-based methods demonstrate consistently strong and robust correlations.

\subsection{Explaining Own Decision}
\begin{figure}
    \centering
    \includegraphics[width=0.75\linewidth]{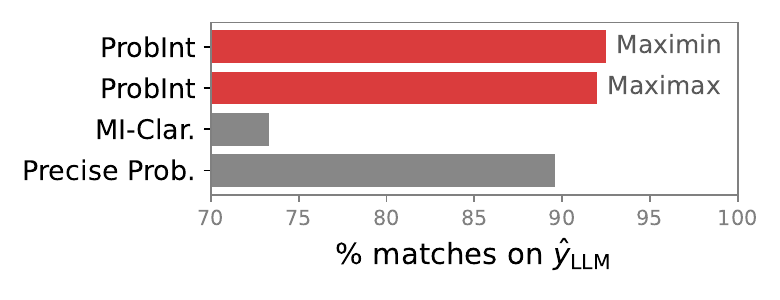}
    \caption{The LLM's prediction $\hat{y}_\text{LLM}$ aligns with the IP-rational decision.}
    \label{fig:decision_belief}
\end{figure}
Previously, we evaluated correctness against the ground-truth $y^\star$. We now instead compare against the model’s own prediction $\hat{y}$. In classical multi-class classification, prediction is obtained via $\hat{y}_\text{precise} = \argmax_{y \in \mathcal{Y}} \hat{p}(y = \text{correct})$, which we refer to as \textsc{Precise Prob}. This rule is fully algorithmic and internally consistent.
In contrast, LLM predictions resemble sampling from a conditional distribution, $\hat{y}_\text{LLM} \sim \hat{p}(y \mid x)$, meaning elicited probabilities need not coincide with the argmax of the predictive distribution. More broadly, this issue relates to faithfulness and self-consistency in LLMs \citep{madsen2024self, matton2025walk}.

\textbf{Bayesian rationality.}
Under uncertainty, a natural self-consistency criterion is Bayesian rationality \citep{harsanyi1978bayesian}, which checks whether $\hat{y}$ matches the argmax of expected utility. To compute utility, we employ \textsc{MI-Clarification} to elicit predictive distributions over answers: $\hat{y}_\text{Bayes} = \argmax_{y \in \mathcal{Y}} \mathbb{E}_{C_i \sim p(C_i \mid x)} \, \hat{p}(y = \text{correct} \mid C_i)$, where $C_i$ denotes clarification contexts.

\textbf{IP-based rationality.}
Decision-making under IP does not yield a single canonical rule as in Bayesian theory. We therefore consider two standard criteria:
\emph{maximin} ($\hat{y}_\text{maximin} = \argmax_{y \in \hat{\mathcal{Y}}} \underline{p}(y)$) and \emph{maximax} ($\hat{y}_\text{maximax} = \argmax_{y \in \hat{\mathcal{Y}}} \overline{p}(y)$). 
We use \textsc{ProbInt} to directly elicit the probability intervals $[\underline{p}(y),\overline{p}(y)]$.

\textbf{Results.}
We evaluate on AmbigQA by measuring the alignment rate between $\hat{y}_\text{LLM}$ and each decision rule. As shown in Figure~\ref{fig:decision_belief}, the maximin and maximax rules exhibit strong alignment with the LLM’s predictions. This suggests that the elicited probability intervals capture uncertainty relevant to answer selection. While this does not imply that LLMs explicitly optimize these rules, it shows that their predictions are compatible with IP-based rationality criteria.

%% file: sections/05_conclusion.tex
\subsection{Discussion}
We provide additional details in the appendix to assess the robustness and efficiency of our approach. First, to justify using approximate MMI instead of exact MMI, we compare the two quantities empirically. The approximation strongly agrees with exact MMI, with Pearson correlations above $0.9$, while producing a speedup of more than $10^6$ (Appendix~\ref{apd:approx_vs_exact}). Second, to rule out prompt-specific artifacts, we conduct a prompt-sensitivity ablation and find that different \textsc{ProbInt} variants lead to consistent conclusions, indicating robustness to prompt formulation (Appendix~\ref{apd:prompt_sensitivity}). Finally, to assess consistency between first- and second-order elicitation, we test whether the first-order probability estimate lies within the second-order probability interval. This holds approximately $70\%$ of the time, which is reasonable given stochastic variation from temperature sampling; adding an explicit containment constraint increases containment to $100\%$ without changing the qualitative conclusions (Appendix~\ref{apd:consistency}).

\section{Conclusion and Limitations}
We propose IP-based methods for higher-order uncertainty elicitation. Across tasks and models, our approach improves elicitation accuracy and internal consistency while remaining cost-efficient. Combined with MMI-based post-processing, IP provides a principled framework for assessing LLM credibility. Overall, our method offers a coherent interface for eliciting, representing, and using uncertainty from black-box models.

Our method shares several limitations with prior work.
First, we assume verbalized uncertainty is approximately rational; although allowing imprecision mitigates, this cannot be fully verified. We also assume the model correctly interprets prompts. Strong classification performance is required—if the model fails to recognize alternative answers, ambiguity cannot be captured. Finally, we focus primarily on Q\&A tasks; extending to settings such as translation or summarization remains future work.